\documentclass{article}

\PassOptionsToPackage{numbers,compress}{natbib}

\usepackage[preprint]{neurips_2026}

\usepackage[utf8]{inputenc} % allow utf-8 input
\usepackage[T1]{fontenc}    % use 8-bit T1 fonts
\usepackage{hyperref}       % hyperlinks
\usepackage{url}            % simple URL typesetting
\usepackage{booktabs}       % professional-quality tables
\usepackage{amsfonts}       % blackboard math symbols
\usepackage{nicefrac}       % compact symbols for 1/2, etc.
\usepackage{microtype}      % microtypography
\usepackage{xcolor}         % colors
\usepackage{tabularx}
\usepackage{graphicx}
\usepackage{amsmath}
\usepackage{amssymb}
\usepackage{bm}
\usepackage{multirow}
\usepackage{multicol}
\usepackage{colortbl}
\usepackage{hhline}
\usepackage{array}
\usepackage{eucal}
\usepackage{wrapfig}
\usepackage{siunitx}
\usepackage{float}
\usepackage[normalem]{ulem}
\usepackage{placeins}
\usepackage{enumitem}

\usepackage{tcolorbox}
\tcbuselibrary{breakable, skins}
\usepackage{capt-of}

\title{Super-Generalist: Towards Comprehensive and Accurate Medical Image Understanding via Generalist–Specialist Synergy}

\author{
\bfseries
Shaoteng~Zhang$^{1,3}$\thanks{Equal contribution.},
Weiwei~Cao$^{1,2}$\footnotemark[1],
Wanxing~Chang$^{1,2}$\footnotemark[1],
Yutong~Xie$^{8}$,
Kai~Cao$^{5}$,
Zaiyi~Liu$^{4}$,
Yu~Shi$^{6}$, \\
\bfseries
Tingbo~Liang$^{7}$,
Qi~Zhang$^{7}$,
Ling~Zhang$^{1,2}$,
Yong~Xia$^{3}$\thanks{Corresponding author: \texttt{yxia@nwpu.edu.cn}, \texttt{jianpeng.zhang0@gmail.com}},
Jianpeng~Zhang$^{1,2}$\footnotemark[2] \\
\\
$^{1}$DAMO Academy, Alibaba Group \quad $^{2}$Hupan Lab \\
$^{3}$Department of Radiology, Ningbo No.\ 2 Hospital, Ningbo, Zhejiang, China \\
$^{4}$Department of Radiology, Guangdong Provincial People’s Hospital, Guangzhou, China \\
$^{5}$Department of Radiology, Shanghai Institution of Pancreatic Disease, Shanghai, China \\
$^{6}$Department of Radiology, Shengjing Hospital of China Medical University, Shenyang, China \\
$^{7}$The First Affiliated Hospital, Zhejiang University School of Medicine, Hangzhou, Zhejiang, China \\
$^{8}$Mohamed bin Zayed University of Artificial Intelligence, Abu Dhabi, United Arab Emirates
}

\begin{document}

\maketitle

\begin{abstract}
Medical images require comprehensive and accurate interpretation to support the diagnosis of diverse clincial conditions. Recent vision–language generalist models offer broad task coverage and promising zero-shot capabilities, yet often lack fine-grained anatomical and lesion awareness for reliable diagnosis and spatial interpretability. In contrast, supervised specialist models achieve strong performance on specific tasks but typically lack generalization across diseases and anatomies.
In this work, we present SuG, a Super-Generalist framework that unifies generalist vision–language learning with specialist objectives, enabling both broad generalization and specialist-level diagnostic capability. We perform specialist-enhanced vision-language alignment in SuG by incorporating spatial priors from multiple segmentation experts, 
including anatomy, class-specific lesion and class-agnostic lesion segmentors that captures lesions beyond anatomies annotated during training. 
To improve lesion grounding capability, we leverage lesion masks as spatial priors to calibrate text-conditioned visual attention, encouraging disease-related semantics to focus on clinically relevant regions.
We evaluate SuG on extensive chest and abdominal CT benchmarks, including CT-RATE, Merlin, MedVL-CT69K, and several in-house tumor datasets. SuG achieves state-of-the-art performance across a wide range of disease diagnosis tasks and surpasses specialist models on several critical tumor diagnosis benchmarks. 
Furthermore, SuG demonstrates strong lesion grounding capability, including robust generalization to lesion types lacking class-specific supervision.
% Code and models will be released upon acceptance.
\end{abstract}

\section{Introduction}
\label{sec:intro}
\begin{figure}
    \centering
    \includegraphics[width=0.68\linewidth]{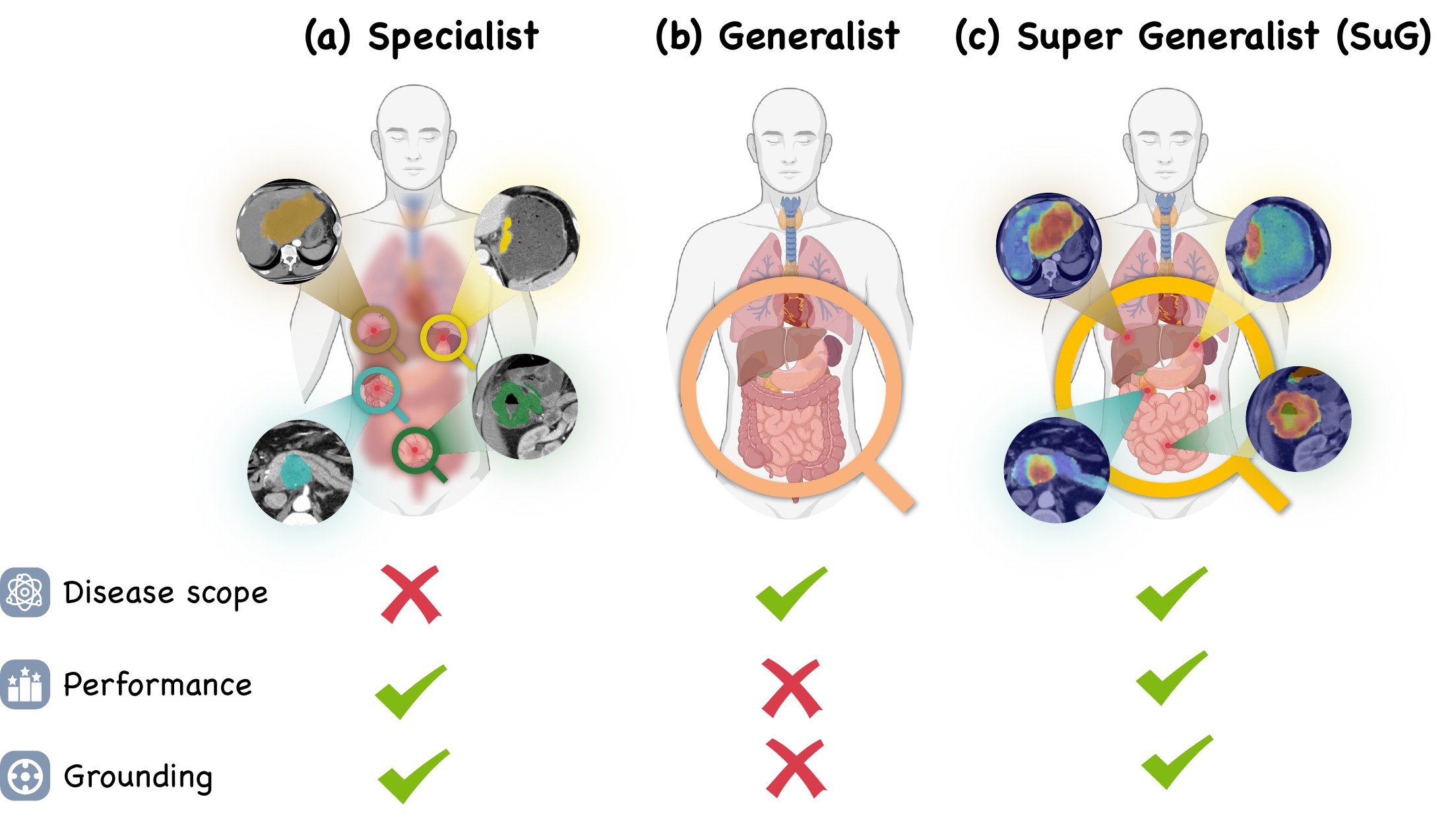}
    \caption{\textbf{Comparison of medical AI paradigms: Specialist, Generalist, and the proposed Super Generalist (SuG).} We evaluate these paradigms across three critical dimensions: \textit{disease scope} (range of tasks), \textit{performance} (diagnostic accuracy), and \textit{grounding} (lesion localization). (a) Specialists excel in performance and grounding but are limited to a narrow disease scope. (b) Conversely, Generalists support a wide scope but suffer from inferior performance and a lack of precise grounding. (c) Our proposed SuG overcomes these trade-offs, achieving high diagnostic accuracy and precise lesion localization across diverse diseases.}
    \label{fig:fig1}
\end{figure}

Toward comprehensive and accurate medical image understanding, a core challenge in modern diagnostic AI is to unify three foundational pillars: broad disease scope, specialist-competitive diagnostic performance, and interpretable lesion grounding~\cite{broad1,broad2,intro-seg1,jin2025chain}.  
A system possessing all three would combine the clinical breadth to handle diverse diseases, the precision to rival human specialists, and the transparency required for clinical decision-making. However, as illustrated in Fig.~\ref{fig:fig1}, current research remains fragmented, with prevailing paradigms forced to sacrifice one of these pillars to excel in another~\cite{intro-seg1, intro-seg2, Cotr, nnUNet, intro-lung, intro-pdac,spec1,spec2,zhang2022contrastive,medvlm1,medvlm2,Merlin}.

Current research is largely split between two dominant approaches: \textit{Specialist} models and \textit{Generalist} models. Specialist models excel in fine-grained performance and spatial grounding but rely heavily on extensive task-specific annotations, limiting their adaptability and scalability across diverse anatomies and clinical scenarios~\cite{intro-seg1, intro-seg2, Cotr, nnUNet, intro-lung, intro-pdac}. 
Conversely, Generalist models utilize vision–language alignment to offer broad diagnostic scope and zero-shot flexibility~\cite{zhang2022contrastive,medvlm1,medvlm2,Merlin}. However, these models struggle to identify fine-grained medical anomalies, as they often rely on global image cues rather than precise lesion features, which limits both their diagnostic accuracy and clinical trustworthiness~\cite{gen-ground1,spe_better1,spe_better2,spe_better3}.

To overcome these limitations, we propose the Super-Generalist (SuG) framework, a novel paradigm designed to harmonize broad disease coverage with specialist-level performance. SuG moves beyond traditional trade-offs by integrating the strengths of both specialists and generalists. By systematically incorporating specialist priors into a generalized vision-language architecture, SuG introduces Generalist–Specialist synergy to achieve broad disease scope, specialist-competitive diagnostic performance, and interpretable lesion grounding. Our approach demonstrates that these capabilities are not mutually exclusive, but can be unified to build a scalable and clinically reliable diagnostic framework for the future of medicine.

SuG serves as a unified architecture that bridges the gap between generalist disease scope and specialist-level performance through three synergistic components:
First, to achieve specialist-competitive performance, we leverage multi-task segmentation objectives with voxel-level annotations across anatomy, class-specific lesion, and class-agnostic lesion targets. This compels the model to encode high-resolution spatial details that are otherwise ignored by conventional generalist architectures.
Second, to extend disease scope and further boost diagnostic performance, we propose a specialist-enhanced vision–language alignment approach. By injecting specialist-derived anatomical and lesion features into generalist embeddings, we enrich cross-modal representations with fine-grained details, significantly strengthening the alignment between visual patterns and clinical text.
Finally, to ensure lesion grounding, we implement an attention calibration strategy that utilizes lesion masks as spatial priors. This steers text-derived queries toward clinically relevant visual tokens, thereby enhancing interpretability and diagnostic traceability.

Overall, SuG achieves state-of-the-art performance in broad-scope disease diagnosis and outperforms specialist models in critical tumor tasks, while demonstrating trustworthy grounding for pan lesions. Our contributions are:
\begin{itemize}[leftmargin=0.3in]
    \item To our best knowledge, we are the first to propose a Generalist–Specialist synergy paradigm for medical image understanding. This paradigm simultaneously achieves broad disease scope, specialist-competitive performance, and interpretable lesion grounding, resolving the inherent trade-offs in existing frameworks.
    \item We propose \textit{Super-Generalist} (SuG), a novel architecture that bridges the gap between wide-scope generalizability and high-performance accuracy by infusing specialist-derived anatomical and lesion priors into generalist vision–language learning.
    \item To achieve robust lesion grounding, we propose a dual-mode lesion segmentor combined with a lesion-guided attention calibration strategy, which anchors diagnostic semantics in clinically relevant regions, ensuring high interpretability.
    \item SuG demonstrates state-of-the-art broad-coverage diagnostic performance and robust clinical interpretability across 10 large-scale CT benchmarks.
\end{itemize}

\begin{figure*}[!htbp]
    \centering
    \includegraphics[width=0.88\linewidth]{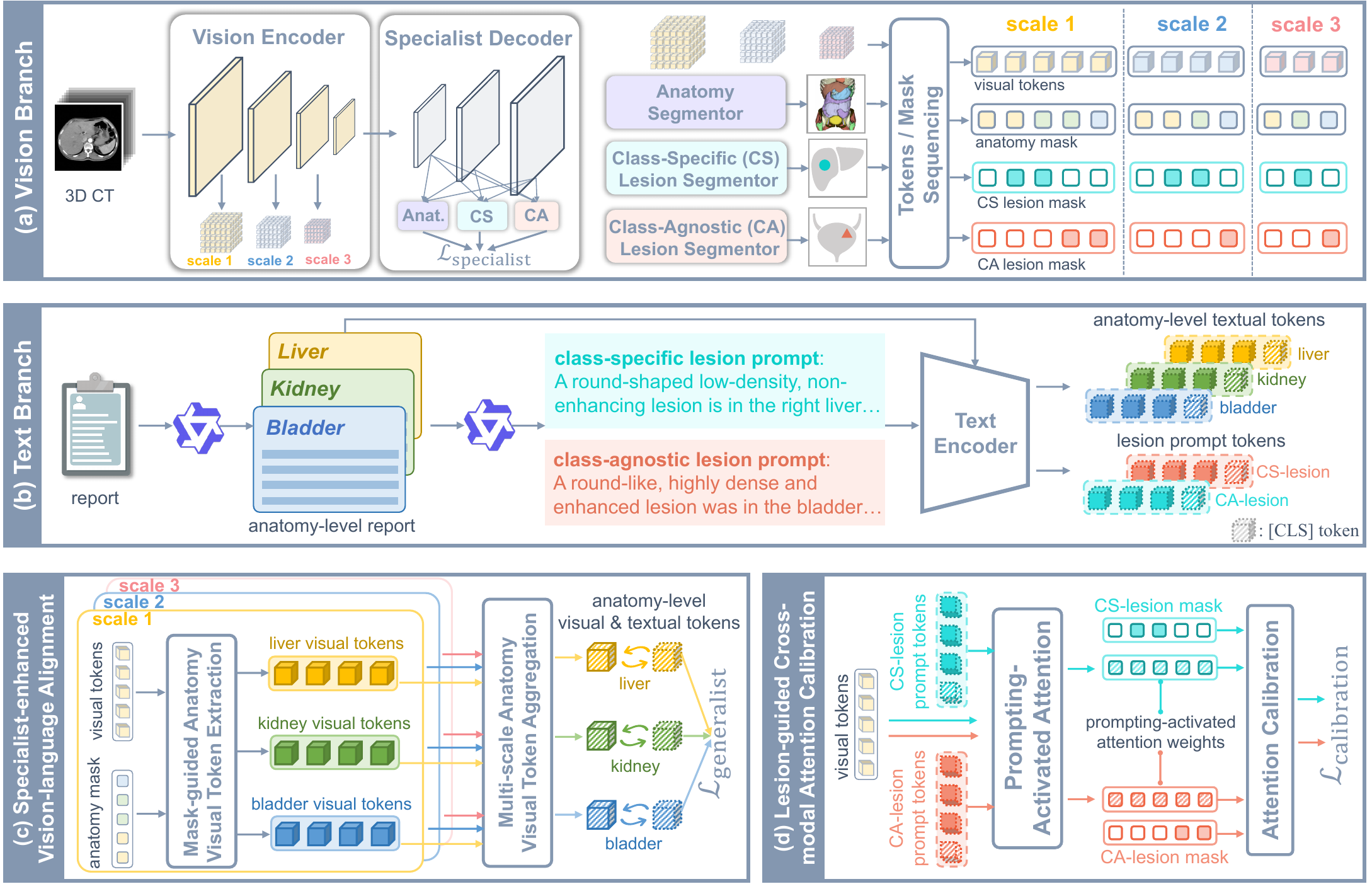}
    \caption{\textbf{An overview of the proposed SuG framework.} SuG consists of two fundamental branches and two synergistic mechanisms: (a) Vision branch extracts multi-scale features and performs anatomy/lesion segmentation via a specialist decoder. 
    (b) Text branch encodes reports into anatomy-level tokens and lesion prompt tokens. 
    (c) Specialist-enhanced vision-language alignment leverages predicted anatomy masks to guide the alignment between multi-scale visual and anatomy-level textual tokens, optimizing the generalist representation. 
    (d) Lesion-guided cross-modal attention calibration dynamically refines text-conditioned visual attention through lesion masks to achieve precise, pathology-aware localization.
    } 
    \label{fig:framework}
\end{figure*}

\section{Method}

\subsection{Overview of SuG}
SuG integrates specialist supervision with generalist vision--language learning to achieve comprehensive and accurate understanding of medical images. As shown in Fig.~\ref{fig:framework}, the framework consists of a vision encoder $\psi$, a specialist decoder $\phi$, and a text encoder $g_{\theta_{\text{text}}}$. The specialist pathway is trained with voxel-level annotations to learn anatomy-aware and lesion-aware visual representations, whereas the generalist pathway is trained with paired images and radiology reports to enable broad cross-modal diagnosis without task-specific labels.

Given a specialist dataset $D_S$ and a generalist dataset $D_G$, SuG is designed to (i) inject anatomical and lesion cues derived from specialist supervision into visual embeddings, (ii) align anatomy-specific image and text representations through contrastive learning, and (iii) calibrate text-conditioned attention with lesion localization priors. Together, these components enable SuG to combine broad diagnostic coverage with fine-grained lesion understanding and grounding.

\subsection{Specialist segmentation across anatomies, lesions, and beyond}
The specialist pathway learns fine-grained anatomical and lesion structures from the voxel-labeled dataset $D_S=\{(I_j,M_j)\}_{j=1}^N$. Given an input CT volume $I_j$, the vision encoder and specialist decoder produce dense features that are processed by three segmentation heads.

\noindent\textbf{Anatomy segmentation.}
An anatomy segmentation head $\omega_{\text{anat}}$ predicts anatomy logits
\begin{equation}
Z^{\mathrm{anat}}_j = f_{\psi,\phi,\omega_{\mathrm{anat}}}(I_j), \qquad
Z^{\mathrm{anat}}_j \in \mathbb{R}^{C_{\mathrm{anat}} \times D \times H \times W},
\end{equation}
where $C_{\mathrm{anat}}$ denotes the number of anatomy classes, including the background class. The anatomy loss is defined as
\begin{equation}
\mathcal{L}_{\mathrm{anat}} = \mathcal{L}_{\mathrm{seg}}\big(Z^{\mathrm{anat}}_j, M^{\mathrm{anat}}_j\big),
\end{equation}
where $\mathcal{L}_{\mathrm{seg}}$ is the standard segmentation loss composed of cross-entropy and Dice terms~\cite{nnUNet}.

\noindent\textbf{Class-specific lesion segmentation.}
For anatomies with lesion annotations, a lesion segmentation head $\omega_{\text{CS}}$ predicts lesion logits
\begin{equation}
Z^{\mathrm{CS}}_j = f_{\psi,\phi,\omega_{\mathrm{CS}}}(I_j), \qquad
Z^{\mathrm{CS}}_j \in \mathbb{R}^{C_{\mathrm{CS}} \times D \times H \times W}.
\end{equation}
Since each sample provides annotation for only one lesion class, we use a binary class indicator $\mathcal{C}_j \in \{0,1\}^{C_{\mathrm{CS}}}$ to mask irrelevant classes:
\begin{equation}
\mathcal{L}_{\mathrm{CS}}=
\mathcal{L}_{\mathrm{seg}}\big((\mathcal{C}_j \odot Z^{\mathrm{CS}}_j),\, (\mathcal{C}_j \odot M^{\mathrm{CS}}_j)\big).
\end{equation}

\noindent\textbf{Class-agnostic lesion segmentation.}
To extend lesion modeling beyond anatomies with lesion labels, we introduce a class-agnostic lesion segmentation head $\omega_{\text{CA}}$:
\begin{equation}
Z^{\text{CA}}_j = f_{\psi,\phi,\omega_{\text{CA}}}(I_j), \qquad
Z^{\text{CA}}_j \in \mathbb{R}^{2 \times D \times H \times W}.
\end{equation}
Its target $M^{\text{CA}}_j$ merges all annotated lesions into a single foreground class. A central challenge is that lesion annotations are anatomy-specific; therefore, voxels outside the annotated anatomies are often unlabeled rather than truly negative. We therefore optimize this branch only within a binary validity mask $\mathcal{V}_j$, which marks regions with trustworthy supervision and excludes unlabeled anatomies from the loss (see Fig.~\ref{app_fig:masked_loss} for an illustration):
\begin{equation}
\mathcal{L}_{\text{CA}} =
\mathcal{L}_{\mathrm{seg}}\big(Z^{\text{CA}}_j \odot \mathcal{V}_j,\,
M^{\text{CA}}_j \odot \mathcal{V}_j\big).
\end{equation}
The overall specialist loss is $\mathcal{L}_{\mathrm{specialist}} = \mathcal{L}_{\mathrm{anat}} + \mathcal{L}_{\mathrm{CS}} + \mathcal{L}_{\mathrm{CA}}.$

\subsection{Specialist-enhanced vision--language alignment}
We next transfer anatomical knowledge learned from specialist supervision to the generalist vision--language alignment process. The main idea is to extract anatomy-specific visual tokens with masks predicted by the specialist pathway, aggregate them across scales, and align them with anatomy-specific textual descriptions extracted from the paired report.

\noindent\textbf{Anatomy vision embedding.}
Let $\mathbf{f}^{(s)}_j$ denote the feature map at scale $s \in \{1,\dots,S\}$. For anatomy category $t$, we use the predicted anatomy mask to select the corresponding tokens:
\[
\mathcal{A}^{(s)}_{j,t} = \{\mathbf{f}^{(s)}_{j,p}\mid p \in \mathcal{R}^{s}_{j,t}\},
\]
where $\mathcal{R}^{s}_{j,t}$ denotes the masked spatial positions. We then average the selected tokens at each scale and normalize the pooled feature:
\[
\bar{\mathbf{a}}^{(s)}_{j,t} = \frac{1}{|\mathcal{R}^{s}_{j,t}|}\sum_{p \in \mathcal{R}^{s}_{j,t}} \mathbf{f}^{(s)}_{j,p}, \qquad
\tilde{\mathbf{a}}^{(s)}_{j,t} = \frac{\bar{\mathbf{a}}^{(s)}_{j,t}}{\|\bar{\mathbf{a}}^{(s)}_{j,t}\|_2}.
\]
The final anatomy-aware visual embedding is obtained by averaging the normalized features across scales:
\[
\mathbf{a}_{j,t} = \frac{1}{S}\sum_{s=1}^{S}\tilde{\mathbf{a}}^{(s)}_{j,t}.
\]
This normalization is important because pooled features at different scales may have different magnitudes; otherwise, large-norm scales would dominate the multi-scale embedding.

\noindent\textbf{Anatomy text embedding.}
For each anatomy category $t$, we extract the corresponding textual description $r_{j,t}$ from report $R_j$ using Qwen LLM~\cite{qwen} (see Appendix Section~\ref{app_sec:organ_prompt}), and encode it with text encoder:
\[
\mathbf{t}_{j,t} = g_{\theta_{\text{text}}}(r_{j,t}).
\]

\noindent\textbf{Anatomy-wise contrastive learning.}
Given a mini-batch, let $T$ denote the number of anatomy categories, and let $N_t$ denote the number of valid samples for anatomy category $t$, i.e., samples with both anatomy-specific visual and textual embeddings. We denote the visual and textual embeddings of the $j$-th valid sample for anatomy category $t$ as $\mathbf{a}_{j,t}$ and $\mathbf{t}_{j,t}$, respectively. For each anatomy category, contrastive matching is performed only within its valid samples. The image-to-text matching probability is
\begin{equation}
p^{i2t}_{j,t,k}
=
\frac{
\exp\big(\mathrm{sim}(\mathbf{a}_{j,t}, \mathbf{t}_{k,t})/\tau\big)
}{
\sum_{k'=1}^{N_t}
\exp\big(\mathrm{sim}(\mathbf{a}_{j,t}, \mathbf{t}_{k',t})/\tau\big)
},
\end{equation}
and the text-to-image probability is defined symmetrically as
\begin{equation}
p^{t2i}_{j,t,k}
=
\frac{
\exp\big(\mathrm{sim}(\mathbf{t}_{j,t}, \mathbf{a}_{k,t})/\tau\big)
}{
\sum_{k'=1}^{N_t}
\exp\big(\mathrm{sim}(\mathbf{t}_{j,t}, \mathbf{a}_{k',t})/\tau\big)
}.
\end{equation}
Let $\mathbf{y}^{i2t}_{j,t}, \mathbf{y}^{t2i}_{j,t}\in\{0,1\}^{N_t}$ denote the one-hot matching labels, where the positive entry corresponds to the paired sample with index $j$. The loss is
\begin{equation}
\mathcal{L}_{\mathrm{generalist}}
=
\frac{1}{2}\sum_{t=1}^{T}
\frac{1}{N_t}\sum_{j=1}^{N_t}
\left[
\mathrm{H}\big(\mathbf{y}^{i2t}_{j,t}, \mathbf{p}^{i2t}_{j,t}\big)
+
\mathrm{H}\big(\mathbf{y}^{t2i}_{j,t}, \mathbf{p}^{t2i}_{j,t}\big)
\right],
\end{equation}
where $\mathrm{sim}(\cdot,\cdot)$ denotes cosine similarity, $\tau$ is a learnable temperature, and $\mathrm{H}(\cdot,\cdot)$ is the cross-entropy loss.

\subsection{Lesion-guided cross-modal attention calibration}
To improve focal lesion diagnosis and localization, we further calibrate text-conditioned attention by using lesion masks from the specialist pathway. Given a lesion-related description $r_j^{\text{lesion}}$ extracted from report $R_j$ (see Appendix Section~\ref{app_sec:core_sign_prompt}), we obtain a lesion prompt embedding $\mathbf{h}_j$.
% \[
% \mathbf{h}_j = g_{\theta_{\text{text}}}(r_j^{\text{lesion}}).
% \]
For a visual feature map $\mathbf{f}^{(s)}_j$ at scale $s$, we compute text-conditioned attention at position $q$ as
\[
\alpha^{(s)}_{j,q} = \sigma\!\left(\mathbf{h}_j \cdot \mathbf{f}^{(s)}_{j,q}\right),
\]
where $\sigma(\cdot)$ denotes the sigmoid function. We supervise this attention map with the resized lesion mask $\mathfrak{R}(\hat{M}^{\text{les}}_j)$:
\begin{equation}
\mathcal{L}^{(s)}_{\mathrm{attn}}
=
-\frac{1}{D_sH_sW_s}\sum_q
\Big[
\mathfrak{R}(\hat{M}^{\text{les}}_j)_q \log \alpha^{(s)}_{j,q}
+
\big(1-\mathfrak{R}(\hat{M}^{\text{les}}_j)_q\big)\log\big(1-\alpha^{(s)}_{j,q}\big)
\Big].
\end{equation}
The final calibration loss  $\mathcal{L}_{\mathrm{calibration}}=\frac{1}{S}\sum_{s=1}^{S}\mathcal{L}^{(s)}_{\mathrm{attn}}$.
To reduce noise, we apply report-conditioned filtering: if a report does not mention a lesion in a given anatomy, the corresponding lesion prediction is excluded from supervision.

\subsection{Progressive multi-stage optimization}
We adopt a three-stage training pipeline to stably integrate specialist segmentation, generalist alignment, and lesion-guided calibration. In Stage 1, only the specialist pathway is trained: $\mathcal{L}_{\mathrm{stage1}}=\mathcal{L}_{\mathrm{specialist}}$.
In Stage 2, training is initialized from Stage 1, and segmentation and anatomy-wise vision--language alignment are jointly optimized: $\mathcal{L}_{\mathrm{stage2}}=\mathcal{L}_{\mathrm{specialist}}+\mathcal{L}_{\mathrm{generalist}}$.
In Stage 3, training is initialized from Stage 2, and lesion-guided attention calibration is further introduced:
$\mathcal{L}_{\mathrm{stage3}}=\mathcal{L}_{\mathrm{specialist}}+\mathcal{L}_{\mathrm{generalist}}+\mathcal{L}_{\mathrm{calibration}}$.
This progressive optimization strategy yields stable convergence while balancing segmentation, cross-modal alignment, and lesion grounding.

\section{Experiments}
\subsection{Experimental setup}
\noindent\textbf{Training and evaluation settings.} We train and evaluate SuG across three experimental settings, as summarized in Table~\ref{tab:exp_settings}. Our assessment covers both abdominal and chest CT scenarios. For abdomen, we utilize two large-scale image--report datasets, MedVL-CT69K~\cite{fvlm} and Merlin~\cite{Merlin}, alongside four anatomy-specific tumor segmentation datasets collectively referred to as LesionSegAbdomen. For chest CT, we employ CT-RATE~\cite{CT-RATE}, the external benchmark RAD-ChestCT~\cite{RADChest}, and a lung tumor segmentation dataset, LesionSegLung. Leveraging these datasets, we evaluate SuG's super-generalist capabilities across three dimensions: (i) general disease diagnosis, (ii) fine-grained lesion-centric diagnosis, (iii) lesion grounding.
Further details on datasets, configurations, and implementation are in Appendix~\ref{app_sec:dataset} and~\ref{app_sec:impl_detail}.

\noindent\textbf{Evaluation protocols.} For disease diagnosis, we follow fVLM~\cite{fvlm} to derive diagnosis scores by applying softmax over the similarities between image embeddings and the text embeddings of paired positive/negative text prompts. For lesion grounding, we generate disease-specific attention maps by calculating the similarity between positive prompt embeddings and image patch tokens. These maps serve as predicted heatmaps for both quantitative assessment and qualitative visualization.

\noindent\textbf{Comparison with state-of-the-art methods.} We evaluate SuG against a broad spectrum of specialist-only (S) and generalist-only (G) methods. 
Additionally, we also benchmark against OpenVocabCT~\cite{OpenVocabCT}, HCFNet~\cite{HCFNet}, and VLWS~\cite{VLWS}, which utilize generalist architectures for specialist diagnostic tasks. Notably, as SuG is the first framework to systematically formalize the specialist-generalist (S+G) synergy paradigm, we adapt these methods to this framework to ensure a fair comparison. This alignment enables all baselines to leverage both lesion-specific priors and general diagnostic knowledge under our unified evaluation protocol. 
Detailed descriptions of these baseline adaptations are provided in Appendix~\ref{app_sec:impl_detail}.

\noindent\textbf{Annotated v.s. unannotated lesions.} To evaluate our diagnosis and grounding performance comprehensively, we categorize lesions into two groups based on the availability of supervision: \textit{annotated} lesions, which possess specific category labels, and \textit{unannotated} lesions, which lack such explicit supervision. This distinction enables a dual assessment of our model: its fine-grained performance on supervised targets and its generalization capability toward unlabeled abnormalities.

\begin{table}[t]
\centering
\small
\caption{\textbf{Overview of the experimental settings.} Each setting integrates a specialist lesion segmentation dataset with a generalist image-report dataset. 
}
\label{tab:exp_settings}
\begin{tabular}{lllll}
    \toprule
    \textbf{Setting} & \textbf{Scenario} & \textbf{Specialist Dataset} & \textbf{Generalist Dataset} & \textbf{Main Results} \\
    \midrule
    Setting 1 & Abdomen CT & LesionSegAbdomen & MedVL-CT69K~\cite{fvlm} & Table~\ref{tab1}, 
    \ref{tab:medvl_multitumor_one_table}, \ref{tab:organ_results} \\
    Setting 2 & Abdomen CT & LesionSegAbdomen & Merlin~\cite{Merlin} & Table~\ref{tab:merlin_summary} 
     \\
    Setting 3 & Chest CT & LesionSegLung & CT-RATE~\cite{CT-RATE} & Table~\ref{tab:zero_shot_ct_rate}
    \\
    \bottomrule
\end{tabular}
\end{table}

\begin{table}[!htbp]
\centering
\footnotesize
\setlength{\tabcolsep}{3.6pt}
\renewcommand{\arraystretch}{1.08}
% \caption{General disease diagnosis results on the MedVL-CT69K test set in terms of AUC. Results are reported for diseases grouped into 15 anatomical structures, as well as the overall average across all 54 diseases (Avg). Full names of anatomical abbreviations are provided in Appendix~\ref{tab:anatomy_abbrev}. ``S'', ``G'', and ``S+G'' denote specialist-only, generalist-only, and specialist-generalist synergy methods, respectively. Detailed per-disease results are provided in Table~\ref{app_tab:medvl_detailed}.}
% changwx: 第一句话太啰嗦，要高度概括
% \caption{\textbf{General disease diagnosis performance (AUC) on the MedVL-CT69K test set.} Results are evaluated across 54 diseases grouped into 15 anatomical structures, with the overall average (Avg) reported. ``S'', ``G'', and ``S+G'' denote specialist-only, generalist-only, and specialist-generalist synergy methods, respectively. \red{Note that [DETAILS ABOUT S+G.]  Done}Note that the compared S+G baselines are not methods originally designed for our super-generalist setting; instead, they are adapted from the original methods to enable specialist--generalist synergy under our evaluation protocol. Detailed baseline adaptation and re-training strategies are provided in Appendix~\ref{app_sec:impl_detail}. Full Sensitivity, Specificity and AUC results are provided in Table~\ref{app_tab:medvl_generalist}. Full abbreviations are in Table~\ref{tab:anatomy_abbrev}, with detailed per-disease results in Table~\ref{app_tab:medvl_detailed}.}
\caption{\textbf{General disease diagnosis performance (AUC) on the MedVL-CT69K test set.} Results are evaluated across 54 diseases grouped into 15 anatomical structures, with the overall average (Avg) reported. ``S'', ``G'', and ``S+G'' denote specialist-only, generalist-only, and specialist-generalist synergy methods, respectively. Full Sensitivity, Specificity and AUC results are provided in Table~\ref{app_tab:medvl_generalist}. Full abbreviations are in Table~\ref{tab:anatomy_abbrev}, with detailed per-disease results in Table~\ref{app_tab:medvl_detailed}.}
\resizebox{\textwidth}{!}{%
\begin{tabular}{l c *{16}{S[table-format=2.1]}}
\toprule
Method & Type 
& {Adr} & {Bla} & {Col} & {Eso} & {GB} & {Hrt} & {Kid} & {Liv}
& {Lng} & {Pan} & {PV} & {SI} & {Spl} & {Sto} & {Sac} & {Avg} \\
\midrule
CLIP~\cite{CLIP}               & G   & 63.2 & 65.1 & 65.8 & 67.3 & 59.5 & 44.1 & 59.9 & 72.4 & 88.3 & 65.0 & 78.6 & 74.5 & 71.1 & 68.6 & 47.0 & 68.4 \\
LOVT~\cite{LOVT}               & G   & 60.6 & 70.9 & 67.5 & 89.3 & 61.2 & 78.3 & 60.2 & 69.3 & 80.9 & 67.8 & 75.5 & 70.5 & 66.1 & 69.1 & 48.9 & 69.4 \\
Imitate~\cite{imitate}         & G   & 60.2 & 74.1 & 68.0 & 95.6 & 62.5 & 70.7 & 59.9 & 69.8 & 86.7 & 64.3 & 80.5 & 77.6 & 71.3 & 66.3 & 29.0 & 70.6 \\
BIUD~\cite{BIUD}               & G   & 63.4 & 81.0 & 70.0 & 62.6 & 64.2 & 62.1 & 63.7 & 79.2 & 72.1 & 76.9 & 82.2 & 75.1 & 72.3 & 66.1 & 43.8 & 71.4 \\
Merlin~\cite{Merlin}           & G   & 60.3 & 76.9 & 69.1 & 49.2 & 61.2 & 72.8 & 64.2 & 80.1 & 78.7 & 73.5 & 85.9 & 78.4 & 72.0 & 69.9 & 48.2 & 71.9 \\
fVLM~\cite{fvlm}               & G   & 65.7 & 84.0 & 80.8 & 98.2 & 64.8 & 85.8 & 74.5 & 82.5 & 82.2 & 85.3 & 96.7 & 82.1 & 82.0 & 74.1 & 87.5 & 81.3 \\
ViSD-Boost~\cite{VISD-Boost}   & G   & 68.5 & 81.2 & 81.9 & 98.1 & 72.6 & 90.5 & 78.5 & 85.9 & \textbf{92.7} & 88.9 & 97.3 & 88.3 & 82.9 & 81.1 & 77.5 & 84.9 \\
\midrule
OpenVocabCT~\cite{OpenVocabCT} & S+G & 68.8 & 84.2 & 84.3 & 96.3 & 85.0 & 81.5 & 84.0 & 90.6 & 68.1 & 90.9 & 95.5 & \textbf{89.8} & 94.2 & 87.4 & \textbf{88.4} & 85.8 \\
HCFNet~\cite{HCFNet}           & S+G & 71.2 & 89.2 & 83.0 & \textbf{99.0} & 83.5 & 85.6 & 84.4 & 89.3 & 89.3 & 90.2 & 97.4 & 88.3 & 92.6 & 76.6 & 58.7 & 86.7 \\
VLWS~\cite{VLWS}               & S+G & 70.1 & 88.9 & 84.1 & 95.3 & \textbf{85.6} & 89.5 & 80.4 & 84.9 & 87.0 & \textbf{90.8} & 97.6 & \textbf{89.8} & 94.1 & 85.2 & 80.0 & 86.7 \\
\midrule
\textbf{SuG (Ours)}             & S+G & \textbf{82.9} & \textbf{92.1} & \textbf{87.0} & 97.2 & 84.9 & \textbf{91.6} & \textbf{87.4} & \textbf{92.2} & 90.9 & 90.4 & \textbf{98.4} & 88.9 & \textbf{95.6} & \textbf{88.4} & 88.2 & \textbf{90.1} \\
\bottomrule
\end{tabular}%
}
\label{tab1}
\end{table}

\begin{table}[t]
\centering
\begin{minipage}[t]{0.38\textwidth}
\centering
% \caption{General disease diagnosis results on the Merlin test set. We report both balanced metrics (AUC$^\ast$, F1$^\ast$), following the official Merlin evaluation protocol, and full-set metrics (AUC, F1), following common practice.}
\caption{\textbf{General disease diagnosis performance on the Merlin test set.} 
% Balanced metrics (AUC$^\ast$, F1$^\ast$) are evaluated following the official Merlin protocol, while full-set metrics (AUC, F1) are reported according to common practice. 
Balanced metrics (AUC$^\ast$, F1$^\ast$) follow the official Merlin protocol; full-set metrics (AUC, F1) follow common practice. 
Detailed per-disease results are in Table~\ref{tab:merlin_detailed} and~\ref{tab:merlin_unbalanced_detailed}.}
\resizebox{\linewidth}{!}{
\begin{tabular}{lcccc}
\toprule
Method & AUC$^\ast$ & F1$^\ast$ & AUC & F1 \\
\midrule
CLIP~\cite{hamamci2024foundation} & 64.3 & 62.7 & 64.4 & 41.7 \\
Merlin~\cite{Merlin} & 82.4 & 74.5 & 81.6 & 55.8 \\
fVLM~\cite{fvlm} & 79.1 & 75.2 & 79.0 & 55.0 \\
ViSD-Boost~\cite{VISD-Boost} & 82.8 & 78.4 & 82.5 & 58.4 \\
OpenVocabCT~\cite{OpenVocabCT} & 82.9 & 78.6 & 83.2 & 59.3 \\
HCFNet~\cite{HCFNet} & 83.2 & 78.7 & 82.9 & 57.2 \\
VLWS~\cite{VLWS} & 82.6 & 77.9 & 82.2 & 57.3 \\
\textbf{SuG (Ours)} & \textbf{85.2} & \textbf{80.8} & \textbf{85.1} & \textbf{60.7} \\
\bottomrule
\end{tabular}
}
\label{tab:merlin_summary}
\end{minipage}
\hfill
\begin{minipage}[t]{0.6\textwidth}
\centering
% \caption{General disease diagnosis results on the chest CT benchmarks CT-RATE and RAD-ChestCT. Results are reported for evaluation on the CT-RATE test set and the RAD-ChestCT dataset. \textsuperscript{\textdagger} indicates variants of CT-CLIP that were fine-tuned with supervised learning.}
\caption{\textbf{General disease diagnosis performance on the chest CT benchmarks: CT-RATE test set and RAD-ChestCT.} 
% Evaluations are conducted on the CT-RATE test set and the RAD-ChestCT dataset.
\textsuperscript{\textdagger} denotes CT-CLIP variants fine-tuned with supervised learning. 
% Detailed per-disease results are in Table~\ref{tab:ctrate_detailed} and~\ref{tab:radchest_detailed}.
Detailed results are in Table~\ref{tab:ctrate_detailed} and~\ref{tab:radchest_detailed}.
}
\resizebox{\linewidth}{!}{
\begin{tabular}{l|cccc|cccc}
\toprule
\multirow{2}{*}{Method}
& \multicolumn{4}{c|}{Internal validation (CT-RATE)}
& \multicolumn{4}{c}{External validation (Rad-ChestCT)} \\
\cmidrule(lr){2-5}\cmidrule(lr){6-9}
& Precision & ACC & F1 & AUC
& Precision & ACC & F1 & AUC \\
\midrule
Random~\cite{hamamci2024foundation}       & 18.0 & 50.2 & 57.0 & 50.5 & 26.5 & 50.0 & 55.5 & 49.6 \\
CT-CLIP~\cite{hamamci2024foundation}      & 32.6 & 66.9 & 70.8 & 73.3 & 34.1 & 59.9 & 64.7 & 63.2 \\
CT-VocabFine\textsuperscript{\textdagger} & 35.6 & 70.4 & 73.8 & 76.0 & 35.6 & 62.1 & 66.8 & 65.7 \\
CT-LiPro\textsuperscript{\textdagger}     & 34.3 & 69.1 & 72.6 & 76.1 & 35.1 & 60.6 & 65.0 & 64.7 \\
BIUD~\cite{BIUD}                          & 33.8 & 68.1 & 71.6 & 71.3 & 33.7 & 60.6 & 65.2 & 62.9 \\
Merlin~\cite{Merlin}                      & 33.7 & 67.2 & 70.9 & 72.8 & 34.8 & 61.9 & 66.3 & 64.4 \\
fVLM~\cite{fvlm}                          & 37.9 & 71.8 & 75.1 & 77.8 & \textbf{37.4} & 64.7 & 68.8 & 68.0 \\
ViSD-Boost~\cite{VISD-Boost}              & 38.7 & 73.1 & 75.9 & 79.0 & 34.2 & 65.2 & 69.3 & 69.4 \\
OpenVocabCT~\cite{OpenVocabCT}            & 40.2 & 74.1 & 76.9 & 79.5 & 33.6 & 61.9 & 66.4 & 66.6 \\
HCFNet~\cite{HCFNet}                      & 40.7 & 73.5 & 76.4 & 80.3 & 35.9 & 67.8 & 71.5 & 71.7 \\
VLWS~\cite{VLWS}                          & 39.6 & 73.0 & 76.0 & 79.5 & 35.3 & 65.5 & 69.7 & 70.7 \\
\textbf{SuG (Ours)}                       & \textbf{40.8} & \textbf{74.4} & \textbf{77.1} & \textbf{81.1} & 36.9 & \textbf{68.1} & \textbf{71.9} & \textbf{73.4} \\
\bottomrule
\end{tabular}
}
\label{tab:zero_shot_ct_rate}
\end{minipage}
\end{table}

\begin{table}[t]
% \caption{Fine-grained lesion-centric diagnosis results in terms of AUC on four tumor categories from MedVL-CT69K and LesionSegAbdomen. ``--'' denotes that the corresponding specialist model does not support that category. Full SE/SP/AUC results are provided in the appendix.}
\caption{\textbf{Fine-grained lesion-centric diagnosis performance (AUC) on four annotated tumor categories across MedVL-CT69K and LesionSegAbdomen.} ``--'' denotes that a specific category is not supported by the corresponding specialist model. 
% Full Sensitivity, Specificity and AUC results are provided in Table~\ref{app_tab:medvl_specialist}.
Detailed results are provided in Table~\ref{app_tab:medvl_specialist}.
}
\label{tab:medvl_multitumor_one_table}
\centering
\footnotesize
\setlength{\tabcolsep}{2.2pt}
\renewcommand{\arraystretch}{1.08}

\resizebox{0.84\textwidth}{!}{%
\begin{tabular}{l c *{10}{c}}
\toprule
\multirow{2}{*}{Method} & \multirow{2}{*}{Type} &
\multicolumn{5}{c}{\textbf{MedVL-CT69K}} &
\multicolumn{5}{c}{\textbf{LesionSegAbdomen}} \\
\cmidrule{3-7}\cmidrule{8-12}
& & Colon & Liver & Pancreas & Stomach & Average
  & Colon & Liver & Pancreas & Stomach & Average \\
\midrule
nnUNet-c~\cite{nnUNet}         & S   & 87.2 & --   & --   & --   & --   & 96.0 & --   & --   & --   & --   \\
nnUNet-l~\cite{nnUNet}         & S   & --   & 79.5 & --   & --   & --   & --   & 97.4 & --   & --   & --   \\
nnUNet-p~\cite{nnUNet}         & S   & --   & --   & 78.6 & --   & --   & --   & --   & 94.0 & --   & --   \\
nnUNet-s~\cite{nnUNet}         & S   & --   & --   & --   & 74.7 & --   & --   & --   & --   & 91.5 & --   \\
nnUNet-fuse~\cite{nnUNet}      & S   & 87.2 & 79.5 & 78.6 & 74.7 & 80.0 & 96.0 & 97.4 & 94.0 & 91.5 & 94.7 \\
\midrule
CLIP~\cite{CLIP}               & G   & 65.9 & 60.0 & 73.0 & 71.6 & 67.6 & 64.6 & 62.5 & 66.4 & 68.9 & 65.6 \\
LOVT~\cite{LOVT}               & G   & 68.1 & 58.4 & 74.9 & 69.4 & 67.7 & 83.9 & 55.2 & 70.2 & 54.9 & 66.1 \\
Imitate~\cite{imitate}         & G   & 63.5 & 60.3 & 71.4 & 71.9 & 66.8 & 66.7 & 48.3 & 71.7 & 67.9 & 63.7 \\
BIUD~\cite{BIUD}               & G   & 66.3 & 59.6 & 77.7 & 73.1 & 69.2 & 73.3 & 65.9 & 66.6 & 57.8 & 65.9 \\
Merlin~\cite{Merlin}           & G   & 62.6 & 63.2 & 74.2 & 66.5 & 66.6 & 52.9 & 55.2 & 64.8 & 53.3 & 56.5 \\
fVLM~\cite{fvlm}               & G   & 73.6 & 62.2 & 85.8 & 75.7 & 74.3 & 82.9 & 65.9 & 77.4 & 75.7 & 75.5 \\
ViSD-Boost~\cite{VISD-Boost}   & G   & 83.8 & 64.5 & 94.7 & 84.2 & 81.8 & 76.9 & 73.6 & 88.7 & 67.1 & 76.6 \\
\midrule
OpenVocabCT~\cite{OpenVocabCT} & S+G & 87.0 & 84.9 & \textbf{97.4} & 89.6 & 89.7 & 81.1 & 90.2 & 90.2 & 85.6 & 86.8 \\
HCFNet~\cite{HCFNet}           & S+G & 86.0 & 75.1 & 94.8 & 89.3 & 86.3 & 91.3 & 79.8 & 93.8 & 90.9 & 89.0 \\
VLWS~\cite{VLWS}               & S+G & 83.3 & 84.0 & 96.4 & 88.6 & 88.1 & 86.0 & 86.8 & 87.5 & 88.0 & 87.0 \\
\midrule
\textbf{SuG (Ours)}                            & S+G & \textbf{93.0} & \textbf{90.6} & 96.1 & \textbf{95.2} & \textbf{93.7}
                               & \textbf{97.3} & \textbf{99.3} & \textbf{95.1} & \textbf{92.7} & \textbf{96.1} \\
\bottomrule
\end{tabular}%
}
\end{table}

\subsection{General disease diagnosis}

\noindent\textbf{Abdominal CT benchmarks.}
Table~\ref{tab1} summarizes the performance on the MedVL-CT69K benchmark. While generalist models provide broad disease coverage, they consistently underperform methods that integrate specialist priors. For instance, the leading generalist baseline, ViSD-Boost~\cite{VISD-Boost}, achieves an average AUC of 84.9\%, whereas adapted S+G methods consistently reach the 85.8\%–86.7\% range. SuG achieves state-of-the-art performance with an average AUC of 90.1\%, surpassing the strongest competitor, VLWS~\cite{VLWS}, by 3.4\%. The gain demonstrates that specialist-enhanced generalist learning can substantially boost diagnostic performance.

Table~\ref{tab:merlin_summary} shows a consistent pattern on the Merlin benchmark. SuG consistently outperforms existing methods, achieving 85.2\% AUC$^\ast$ / 80.8\% F1$^\ast$ under the official balanced protocol, and 85.1\% AUC / 60.7\% F1 on the full test set. Notably, MedVL-CT69K and Merlin are derived from distinct patient populations, roughly corresponding to Eastern and Western cohorts, respectively. The consistently superior performance of SuG across these heterogeneous datasets underscores the model’s robust cross-population generalization and the stability of our specialist-enhanced generalist architecture.

\noindent\textbf{Chest CT benchmarks.}
Table~\ref{tab:zero_shot_ct_rate} further evaluates SuG on CT-RATE and the external RAD-ChestCT benchmark. SuG consistently achieves superior performance, attaining 81.1\% AUC on CT-RATE and 73.4\% AUC on RAD-ChestCT.  Compared with the strongest competitor, HCFNet~\cite{HCFNet}, our method yields performance gains of 0.8\% and 1.7\%, respectively. The more pronounced improvement on RAD-ChestCT, an external validation set, highlights SuG's enhanced robustness against cross-dataset distribution shifts. Together with the abdominal results, these findings demonstrate that SuG generalizes reliably across diverse anatomies, populations, and imaging distributions.

\subsection{Fine-grained lesion-centric diagnosis}
Table~\ref{tab:medvl_multitumor_one_table} summarizes the diagnostic performance on four specific tumor categories, evaluated across both MedVL-CT69K and LesionSegAbdomen. This task poses a significant challenge for generalist models, as it necessitates precise lesion perception and localized reasoning.
On MedVL-CT69K, specialist-based nnUNet baselines consistently outperform most pure generalist models, underscoring the critical role of lesion-aware supervision. While existing specialist-generalist synergy methods mitigate this gap, SuG achieves superior performance across all categories by integrating specialist-enhanced alignment with lesion-guided attention calibration.
On LesionSegAbdomen, SuG further improves upon the specialist-level benchmark (nnUNet-fuse, 94.7\% AUC), achieving 96.1\% AUC. These results demonstrate that SuG not only preserves broad diagnostic capabilities but also achieves specialist-level, and even specialist-surpassing, performance in fine-grained lesion-centric diagnosis.

\begin{table}[ht]
% \caption{Quantitative lesion grounding evaluation on four seen tumor types on LesionSegAbdomen test set. 
% % We report voxel-level AUC, AUPR, and Cohen's d($\uparrow$) between the predicted attention maps and lesion masks, where higher values indicate better lesion-background separability and localization quality.
% }
\caption{\textbf{Quantitative lesion grounding evaluation on the LesionSegAbdomen test set.} Results are reported across four annotated tumor types using AUC, AUPR, and Cohen's $d$. These metrics assess the voxel-level correspondence between the predicted attention maps and ground-truth lesion masks.
% Results are reported across four annotated tumor types using three voxel-level metrics. 
% \red{AUC evaluates overall lesion--background discriminability across thresholds, AUPR reflects lesion localization quality under class imbalance, and Cohen's $d$ measures the standardized separation between heatmap responses on lesion and non-lesion voxels. Higher values indicate better localization and clearer lesion--background separation.}
% Results are reported across four seen tumor types using AUC, AUPR, and Cohen's $d$ metrics. \red{AUC pixel-wise}
% We report voxel-level AUC, AUPR, and Cohen's d($\uparrow$) between the predicted attention maps and lesion masks, where higher values indicate better lesion-background separability and localization quality.
}
\label{tab:organ_results}
\centering
\setlength{\tabcolsep}{6pt}
\renewcommand{\arraystretch}{1.2}
\resizebox{\textwidth}{!}{
\begin{tabular}{lccc ccc ccc ccc ccc}
\toprule
\multirow{2}{*}{Method} &
\multicolumn{3}{c}{Colon} &
\multicolumn{3}{c}{Liver} &
\multicolumn{3}{c}{Stomach} &
\multicolumn{3}{c}{Pancreas} &
\multicolumn{3}{c}{Average} \\
\cmidrule(lr){2-4}\cmidrule(lr){5-7}\cmidrule(lr){8-10}\cmidrule(lr){11-13}\cmidrule(lr){14-16}
& AUC & AUPR & Cohen's $d$($\uparrow$)
& AUC & AUPR & Cohen's $d$($\uparrow$)
& AUC & AUPR & Cohen's $d$($\uparrow$)
& AUC & AUPR & Cohen's $d$($\uparrow$)
& AUC & AUPR & Cohen's $d$($\uparrow$) \\
\midrule
Generalist baseline       & 53.5 &  8.1 &  0.1 & 45.1 &  9.6 & -0.2 & 39.5 &  7.2 & -0.4 & 49.7 & 27.6 &  0.0 & 47.0 & 13.1 & -0.1 \\
+ Specialist & 74.0 & 19.5 &  0.8 & 58.6 & 14.6 &  0.3 & 60.9 & 14.4 &  0.4 & 51.2 & 29.5 & -0.1 & 61.2 & 19.5 &  0.4 \\
SuG            & \textbf{92.0} & \textbf{52.8} &  \textbf{2.8} & \textbf{93.5} & \textbf{63.9} &  \textbf{3.8} & \textbf{87.1} & \textbf{47.6} &  \textbf{2.5} & \textbf{90.5} & \textbf{73.8} &  \textbf{2.6} & \textbf{90.8} & \textbf{59.5} &  \textbf{2.9} \\
\bottomrule
\end{tabular}
}
\end{table}

\begin{figure*}[!htbp]
    \centering
    \includegraphics[width=0.84\linewidth]{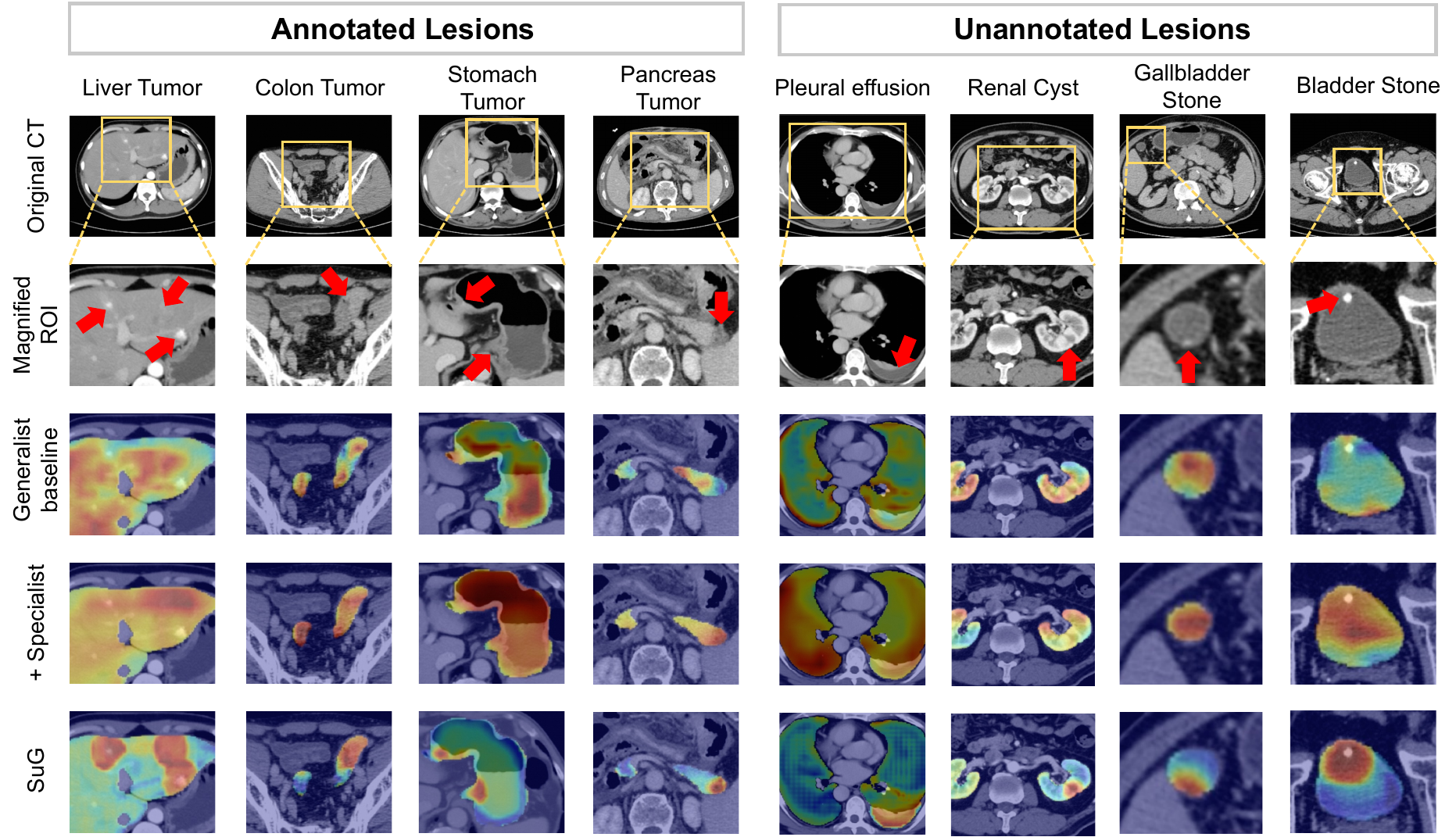}
    \caption{\textbf{Attention visualization on \textit{annotated} and \textit{unannotated} lesions.} The left and right panels display representative cases of four annotated (liver, colon, stomach, pancreas) and unannotated (pleural effusion, renal cyst, gallbladder stone, bladder stone) lesion categories, respectively. Yellow boxes highlight ROIs, while red arrows pinpoint lesion locations. Attention maps from our generalist baseline, the specialist-enhanced generalist model, and SuG are overlaid on CT images. 
    }
    \label{fig:attn}
\end{figure*}

\subsection{Lesion grounding}
\label{sec: lesion grounding}

To facilitate a clear comparison, we compare three incremental configurations. The \textit{Generalist baseline} configuration is a pure generalist vision-language model that shares the same encoders as SuG but lacks specialist supervision and lesion-guided attention calibration. The \textit{+Specialist} configuration further incorporates the specialist branch and enhanced cross-modal alignment, yet still omits lesion-guided attention calibration. Finally, \textit{SuG} denotes the full model integrated with all proposed components. 

\textbf{Annotated lesion grounding.}
Table~\ref{tab:organ_results} reports quantitative voxel-level grounding performance across four annotated lesion categories.
To interpret these metrics, we utilize AUC to assess global lesion–background discriminability, AUPR for localization precision under class imbalance, and Cohen's $d$ to quantify the separation of lesion-to-non-lesion heatmap responses. Higher scores consistently indicate superior localization accuracy and contrast.
A vision-language-based generalist baseline exhibits poor grounding capability, achieving an average AUC of 47.0\%, AUPR of 13.1\%, and Cohen's $d$ of -0.1. 
While integrating specialist supervision provides modest gains, SuG achieves superior grounding with 90.8\% AUC, 59.5\% AUPR, and Cohen’s $d$ of 2.9, demonstrating substantially more faithful, lesion-focused grounding. 

\textbf{Qualitative analysis.}
Fig.~\ref{fig:attn} visualizes attention maps for annotated and unannotated lesions. Compared with the baseline and the specialist-enhanced generalist without calibration, SuG yields sparser, highly localized attention responses while effectively suppressing irrelevant regions. Taking stomach tumors as an illustrative case, the baseline tends to fixate on gastric contents rather than the wall. Although specialist supervision shifts some attention to the gastric wall, it fails to fully decouple from the contents. In contrast, SuG precisely focuses on the thickened gastric wall corresponding to the lesion, effectively filtering out irrelevant background noise.

\textbf{Generalization to unannotated lesions.}
The right panel of Fig.~\ref{fig:attn} further demonstrates that SuG effectively localizes diverse lesions, including pleural effusion, renal cyst, gallbladder stone, and bladder stone, without direct voxel-level supervision for these categories. This suggests that SuG acquires transferable lesion-aware representations, moving beyond simple memorization of the training distribution. Additional visualizations and transfer results are in Fig.~\ref{app_fig:pan_seg}.

\subsection{Ablation study}
Table~\ref{tab:ablation} evaluates the contribution of each component within SuG on the MedVL-CT69K validation set spanning three diagnostic paradigms: general disease diagnosis, lesion-centric diagnosis, and generalization to unannotated localized lesion types (details in Table~\ref{app_tab:val_disease_lists}).

% \begin{table}[t]
% \centering
% \renewcommand{\arraystretch}{1.15}
% \setlength{\tabcolsep}{3pt}

% \resizebox{0.5\textwidth}{!}{%
% \begin{tabular}{lccc}
% \toprule
% Variant & \shortstack{General\\diseases} & \shortstack{Four lesion-\\centric diseases} & \shortstack{Unseen localized\\lesion diseases} \\
% \midrule
% Baseline                    & 80.2 & 75.1 & 74.9 \\
% + Specialist                & 84.7 & 81.6 & 81.7 \\
% + Multi-scale feat.         & 85.7 & 82.3 & 82.0 \\
% + L2 norm                   & 87.4 & 84.4 & 84.8 \\
% + Seen-lesion attn.         & 87.3 & 87.1 & 86.1 \\
% + Pan-lesion attn.          & 89.5 & 87.6 & 89.2 \\
% \bottomrule
% \end{tabular}%
% }
% \caption{Ablation study on the MedVL-CT69K validation set. We progressively incorporate each proposed component into the baseline and report AUC for general disease diagnosis (all 36 diseases), the four lesion-centric diseases, and unseen localized lesion diseases.}
% \label{tab:ablation}
% \end{table}

\begin{wraptable}{r}{0.54\textwidth}
% \vspace{-1.0em}
\centering
% \captionsetup{font=small}
% \caption{Ablation study on the MedVL-CT69K validation set.}
\caption{\textbf{Ablation study (diagnostic AUC) on the MedVL-CT69K validation set.} The effectiveness of each proposed component is evaluated by cumulative integration into the baseline.}
\label{tab:ablation}
\renewcommand{\arraystretch}{1.12}
\setlength{\tabcolsep}{3pt}

\resizebox{0.52\textwidth}{!}{%
\begin{tabular}{lccc}
\toprule
Variant & \shortstack{general\\diseases} & \shortstack{four annotated\\lesion-centric diseases} & \shortstack{unannotated localized \\lesion diseases} \\
\midrule
Generalist baseline            & 80.2 & 75.1 & 74.9 \\
+ Specialist        & 84.7 & 81.6 & 81.7 \\
+ Multi-scale feat. & 85.7 & 82.3 & 82.0 \\
+ $\ell_2$ norm           & 87.4 & 84.4 & 84.8 \\
+ Class-specific lesion attn. & 87.3 & 87.1 & 86.1 \\
+ Class-agnostic lesion attn.  & \textbf{89.5} & \textbf{87.6} & \textbf{89.2} \\
\bottomrule
\end{tabular}%
}
% \vspace{-1.0em}
\end{wraptable}

Building upon the configurations defined in Section~\ref{sec: lesion grounding}, we start from the \textit{Generalist baseline} and incorporate the specialist branch, which yields substantial gains and confirms that specialist supervision provides strong lesion-aware priors. 
Introducing multi-scale feature aggregation and $\ell_2$ normalization further enhances performance, demonstrating the benefits of stable multi-scale fusion for both general and lesion-centric diagnosis. Finally, lesion-guided attention calibration introduces the most significant improvements in lesion-sensitive tasks: class-specific lesion attention calibration optimizes performance on supervised categories, while class-agnostic lesion attention calibration substantially strengthens generalization to unannotated localized lesions. Collectively, these results validate our holistic SuG design and demonstrate that each component contributes in a mutually complementary manner.

\section{Conclusion}
We presented SuG, a unified framework toward super-generalist medical image understanding. By integrating specialist-derived anatomical and lesion-aware priors into vision--language learning and calibrating cross-modal attention with lesion masks, SuG combines broad diagnostic coverage with fine-grained lesion-centric understanding and reliable grounding.
Extensive experiments on abdominal and chest CT benchmarks demonstrate that SuG improves general disease diagnosis, achieves specialist-competitive performance on fine-grained lesion-centric diagnosis, and provides more faithful lesion grounding. These findings support the promise of transferring specialist knowledge into generalist medical AI.

\textbf{Limitations and future work.} First, while currently focused on CT benchmarks, the proposed framework can be extended to accommodate diverse imaging modalities. 
Furthermore, we anticipate that the class-agnostic lesion grounding can be further improved and validated as larger-scale, high-fidelity voxel-level annotations become available.

\begin{ack}
This work was supported by the Zhejiang Province Postdoctoral Research Excellence Funding Program (ZJ2024032).
\end{ack}

\bibliographystyle{unsrtnat}
\bibliography{ref.bib}

\clearpage
\appendix

\setcounter{figure}{0}
\setcounter{table}{0}
\setcounter{equation}{0}

\renewcommand{\thefigure}{A\arabic{figure}}
\renewcommand{\thetable}{A\arabic{table}}
\renewcommand{\theequation}{A\arabic{equation}}

\renewcommand{\theHfigure}{A\arabic{figure}}
\renewcommand{\theHtable}{A\arabic{table}}
\renewcommand{\theHequation}{A\arabic{equation}}

\section{Related work}

\subsection{Medical specialist AI}
Specialist models have long been the dominant paradigm in medical image analysis, typically designed for disease‑specific supervised learning tasks such as segmentation~\cite{medseg1,medseg2,medseg3}, detection~\cite{meddet1,meddet2, meddet3}, grading\cite{grading1,grading2,grading3}, and prognosis prediction~\cite{prognosis1,prognosis2,prognosis3}. Among these tasks, segmentation and localization of anatomical structures and lesions are fundamental challenges. U‑Net~\cite{Unet} and its variants~\cite{nnFormer,nnUNet,Cotr,SwinUNETR} form the foundation of most specialist models. In particular, nnUNet~\cite{nnUNet} has become one of the most successful examples, achieving highly competitive performance on a wide range of medical image segmentation benchmarks.
However, developing specialist models inherently requires high‑quality annotated datasets. Taking segmentation as an example, training such models typically demands large‑scale voxel‑level annotations. While specialist models excel in their dedicated tasks and can achieve very high accuracy, they face two key limitations. First, extending to new tasks incurs high costs, as it often requires collecting new annotations and retraining from scratch. Second, their task scope is narrow, with limited cross‑anatomy and cross‑disease transferability in open‑world settings.

\subsection{Medical generalist AI}
Vision–language models (VLMs) have emerged as a promising pathway toward generalist medical AI. A mainstream approach follows the CLIP-style contrastive learning paradigm, where paired medical images and radiology reports are used to align visual and textual representations~\cite{zhang2022contrastive,medvlm1,medvlm2,medvlm3,medvlm4,medvlm5,medvlm6}. Once trained, such models exhibit broad task generalization capabilities, enabling zero-shot or few-shot diagnosis across diverse conditions.
Due to the limited availability of large-scale paired medical data, early explorations focused primarily on 2D chest X‑ray scenarios. Pioneer studies~\cite{zhang2022contrastive,tiu2022expert} demonstrated that this paradigm can endow models with medical image understanding and diagnostic capabilities without additional annotated labels, leveraging only image–report pairs.
To improve cross-modal alignment quality, subsequent methods have incorporated local alignment into contrastive learning. For example, LoVT~\cite{LOVT} and MGCA~\cite{MGCA} explicitly model correspondences between local image regions and corresponding report phrases or sentences. This alleviates the semantic loss caused by purely global alignment, capturing fine-grained clinical details.
More recently, researchers have extended medical VLMs to 3D volumetric data such as CT, exploring generalist AI in 3D medical image understanding~\cite{Merlin,hamamci2024foundation,fvlm,VISD-Boost}. 
Although GSCo~\cite{he2026generalizable} proposes a collaborative framework for generalist and specialist models, it limits the specialist’s role to an external advisor that provides inference-time guidance, rather than achieving a deep architectural integration. In contrast, our approach internalizes specialist knowledge directly within the model architecture. This structural synergy allows us to simultaneously satisfy three core objectives, i.e. broad diagnostic coverage, specialist-competitive accuracy, and interpretable lesion grounding, which remain fundamentally different from GSCo’s retrieve-and-predict paradigm.
As a result, these generalist approaches still remain behind specialist models on tasks that require fine-grained lesion perception and localization, particularly when interpretable lesion-grounded diagnosis is needed. 

\subsection{Language-driven medical image segmentation}
Another related line of research explores the use of pretrained vision--language alignment knowledge for medical image segmentation. These methods are closely related to our setting at the architectural level, as they also integrate visual encoding, textual encoding, and dense prediction within a unified framework. However, their objective is fundamentally different from ours. Existing language-driven segmentation methods are primarily designed to exploit such cross-modal alignment to improve segmentation performance, i.e., to build a stronger specialist model. Their main emphasis is typically on how to better transfer pretrained VLM knowledge to segmentation, for example by designing decoder architectures that more effectively fuse image and text features to enhance segmentation quality~\cite{OpenVocabCT,HCFNet,VLWS}.
In contrast, our goal is not to use language supervision mainly to improve segmentation itself. Instead, we use segmentation-derived specialist knowledge to retain the broad diagnostic capability of a generalist model while improving its ability to recognize and localize focal lesions.

\begin{figure}[!htbp]
    \centering
    \includegraphics[width=0.50\linewidth]{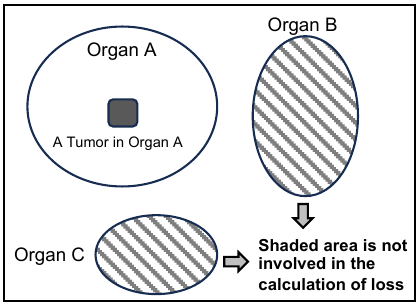}
    % \caption{Masked loss strategy for class-agnostic pan-lesion segmentation.}
    \caption{\textbf{Illustration of the masked loss strategy for class-agnostic lesion segmentation.}}
    \label{app_fig:masked_loss}
\end{figure}

\section{Prompt Templates for Report Parsing}
\label{app_sec:prompts}

\subsection{Prompt template for anatomy-level report description extraction}
\label{app_sec:organ_prompt}

For reproducibility, we provide the prompt template used to extract anatomy-level textual descriptions from CT reports using Qwen. These extracted descriptions are used to construct the anatomy-specific text input.

In the template, \texttt{\{TARGET\_ANATOMY\}} denotes the queried anatomical region, and \texttt{\{REPORT\_TEXT\}} denotes the input report.

\begin{figure*}[t]
\centering
\fbox{
\parbox{0.96\textwidth}{
\footnotesize
\ttfamily

You are a professional radiologist. Please extract the descriptive information about the specific anatomy from this English CT diagnostic report.

\medskip
\textbf{Target anatomy:} \{TARGET\_ANATOMY\}

\medskip
\textbf{Requirements:}
\begin{itemize}
    \item Extract only the content directly related to the target anatomy.
    \item If the report mentions specific subregions, parts, involved sites, or laterality of the target anatomy, preserve these anatomical details in the extracted description.
    \item Keep the extracted text concise and faithful to the original report.
    \item Do not add explanations, background knowledge, or inferred findings that are not explicitly stated in the report.
    \item Even if the target anatomy has multiple independent parts, treat it as a whole and return only one integrated description.
\end{itemize}
\textbf{Output format:}
\begin{itemize}
    \item Output exactly one line in the format: \texttt{\{TARGET\_ANATOMY\}: description}.
    \item Always use the queried target anatomy as the unified prefix.
    \item If no relevant information is found, output: \texttt{\{TARGET\_ANATOMY\}: no relevant finding}.
    \item Do not return multiple entries, JSON, or additional comments.
\end{itemize}

\textbf{CT report input:} \{REPORT\_TEXT\}
}}
\caption{\textbf{Prompt template for anatomy-specific report extraction.}
We query the LLM independently for each target anatomy to ensure high consistency and quality in the extracted descriptions, which also simplifies the parsing process compared to multi-anatomy extraction.}
\label{fig:organ_prompt}
\end{figure*}
\subsection{Prompt template for disease and core imaging sign extraction}
\label{app_sec:core_sign_prompt}

We also provide the prompt used to extract diseases and their supporting core imaging signs from CT reports. This prompt is used to derive lesion-related descriptions for attention calibration and to support report-conditioned filtering.

In the template, \texttt{\{TARGET\_ANATOMY\_LIST\}} refers to the predefined target anatomy list (see Table~\ref{app_tab:target_organs}), and \texttt{\{REPORT\_TEXT\}} denotes the input report.

\begin{figure*}[t]
\centering
\fbox{
\parbox{0.96\textwidth}{
\footnotesize
\texttt{
You are a medical imaging report analysis assistant. Extract diseases and their supporting core imaging signs from an English CT report.\\
\\
Target anatomies: \{TARGET\_ANATOMY\_LIST\}\\
\\
Requirements:\\
- Each disease must be mapped to exactly one target anatomy.\\
- Discard diseases that cannot be clearly mapped to the target anatomy list.\\
- Output keys must follow: anatomy-english\_disease\_name-lesion\_type.\\
- If IMPRESSION / CONCLUSION / ASSESSMENT exists, use it as the primary diagnosis list.\\
- Otherwise, convert abnormal FINDINGS statements into disease/problem labels.\\
\\
Lesion types:\\
- Localized Lesion: focal abnormality, e.g., lesion, mass, nodule, tumor, cyst, calcification, thrombus, aneurysm, stricture, focal wall thickening.\\
- Diffuse Lesion: generalized or anatomy-wide abnormality, e.g., diffuse inflammation, steatosis, hepatomegaly, edema, generalized dilatation.\\
- If uncertain, default to Diffuse Lesion.\\
\\
Procedure:\\
1. Extract explicit abnormal imaging findings from FINDINGS/DESCRIPTION.\\
2. Keep only generic core signs; remove size, exact location, laterality, measurements, and comparison terms.\\
3. Identify diseases from IMPRESSION/CONCLUSION if present; otherwise derive disease/problem labels from abnormal FINDINGS.\\
4. For each disease, collect the supporting core signs from FINDINGS.\\
5. Map each disease to exactly one target anatomy using predefined rules and standard medical knowledge.\\
6. Build a comma-separated English core sign string.\\
\\
Representative anatomy mapping rules:\\
- large bowel: colon, rectum, cecum, appendix, anal canal\\
- small bowel: jejunum, ileum\\
- duodenum: duodenum only\\
- gallbladder: gallbladder, cystic duct, and biliary duct terms if no separate duct anatomy exists\\
- portal vein: portal vein system, including splenic vein\\
- Discard anatomies not in the target list, e.g., uterus, ovaries, prostate, lymph nodes\\
\\
Output format:\\
- Output only a single JSON dictionary, with no extra text.\\
- Key format: anatomy-english\_disease\_name-lesion\_type\\
- lesion\_type must be exactly ``Localized Lesion'' or ``Diffuse Lesion''\\
- Value: the English core sign string (comma-separated).\\
\\
CT report input: \{REPORT\_TEXT\}
}
}}
\caption{\textbf{Prompt template designed for disease and core imaging sign extraction from CT reports.}}
\label{fig:core_sign_prompt}
\end{figure*}

\begin{table*}[t]
\centering
\caption{\textbf{List of anatomical structures used in prompts for disease and core imaging sign extraction.}}
\label{app_tab:target_organs}
\begin{tabular}{lll}
\toprule
adrenal gland & aorta & erector spinae muscle \\
brain & clavicle & large bowel \\
duodenum & esophagus & face \\
femur & gallbladder & gluteus muscle \\
heart & hip joint & humerus \\
iliac artery & iliac vena & iliopsoas muscle \\
inferior vena cava & kidney & liver \\
lung & pancreas & portal vein \\
pulmonary artery & rib & sacrum \\
scapula & small bowel & spleen \\
stomach & trachea & bladder \\
cervical vertebrae & lumbar vertebrae & thoracic vertebrae \\
\bottomrule
\end{tabular}
\end{table*}

\section{Experiment setting Details}
\label{app_sec:experiment_setting}

\subsection{Dataset}
\label{app_sec:dataset}
\subsubsection{Abdomen scenario}
For the abdominal setting, we use two large-scale image–report datasets, MedVL-CT69K~\cite{fvlm} and Merlin~\cite{Merlin}, together with four anatomy-specific tumor segmentation datasets, which we collectively denote as LesionSegAbdomen. The image–report datasets contain abdominal CT scans paired with detailed radiology reports and are mainly used to support generalist vision–language learning and broad-coverage diagnosis. In contrast, LesionSegAbdomen provides voxel-level lesion annotations and is used for specialist supervision as well as lesion-centric evaluation.

\textbf{MedVL-CT69K}. MedVL-CT69K~\cite{fvlm} is a large-scale abdominal CT dataset containing 272,124 CT scans from 69,086 patients, each paired with a corresponding radiology report. Its validation and test sets are annotated by expert radiologists with 36 and 54 disease labels, respectively. Following the official split in the fVLM~\cite{fvlm}, we divide the dataset into training, validation, and test sets with 64,476, 1,151, and 3,459 patients, respectively.

\textbf{Merlin}. Merlin~\cite{Merlin} is a public abdominal CT dataset collected from the Stanford Hospital Emergency Department between 2012 and 2018. It contains 25,494 CT scans from 18,317 patients, each paired with a radiology report. Following the original benchmark setting~\cite{Merlin}, we use 11,010 patients for training, 3644 patients for validation, and 3,667 patients for testing. The dataset provides ground-truth labels for 30 disease categories, which we use to evaluate diagnostic performance.

\textbf{LesionSegAbdomen}. We refer to the union of four anatomy-specific abdominal tumor segmentation datasets as LesionSegAbdomen. It includes LesionSegC (colorectal), LesionSegS (stomach), LesionSegL (liver), and LesionSegP (pancreas), each providing voxel-level lesion annotations for its corresponding anatomy. In total, these four datasets contain 23,990 CT scans from 8,010 patients. We perform train and test splitting independently for each sub-dataset, resulting in 6,851 patients for training and 1,159 patients for testing after aggregation (see Table~\ref{app_tab:tumorseg_details}). Each CT scan is associated with a tumor mask annotated by senior radiology experts.

\subsubsection{Chest scenario}
For the chest setting, we use three CT datasets to evaluate the generalization of SuG to chest imaging: the large-scale image–report dataset CT-RATE~\cite{CT-RATE}, the external multi-disease benchmark RAD-ChestCT~\cite{RADChest}, and a lung tumor segmentation dataset, denoted as LesionSegLung, for specialist supervision.

\textbf{CT-RATE and RAD-ChestCT}. CT-RATE~\cite{CT-RATE} contains 25,692 non-contrast chest CT volumes from 21,304 unique patients, each paired with a radiology report. Following the benchmark protocol, we use 20,000 patients for training and 1,304 patients for testing. To assess external generalization, we further evaluate on RAD-ChestCT dataset~\cite{RADChest}, which contains 3,630 chest CT scans. Following prior work~\cite{fvlm}, we report diagnostic performance on 16 diseases for CT-RATE and 13 diseases for RAD-ChestCT. 

\textbf{LesionSegLung}. For specialist supervision in the chest setting, we collect a lung tumor segmentation dataset, LesionSegLung. It contains 1,883 patients, randomly split into 1,783 training cases and 100 test cases (see Table~\ref{app_tab:tumorseg_details}). Each CT scan is annotated with a lesion mask by senior radiology experts. 

\begin{table*}[t]
\centering
\caption{\textbf{Statistics of lesion segmentation datasets.} Listed are the counts of CT scans and patients across train, test, and total sets for each dataset.}
\label{app_tab:tumorseg_details}
\renewcommand{\arraystretch}{1.1}
\setlength{\tabcolsep}{6pt}
\resizebox{\textwidth}{!}{
\begin{tabular}{llcccccc}
\toprule
\multirow{2}{*}{Scenario} & \multirow{2}{*}{Dataset} & \multicolumn{2}{c}{Train} & \multicolumn{2}{c}{Test} & \multicolumn{2}{c}{Total} \\
\cmidrule(lr){3-4} \cmidrule(lr){5-6} \cmidrule(lr){7-8}
& & CT num. & Patient num. & CT num. & Patient num. & CT num. & Patient num. \\
\midrule
\multirow{4}{*}{Abdomen}
& LesionSegC & 3385 & 847  & 435  & 100 & 3820 & 947  \\
& LesionSegS & 3462 & 1865 & 178  & 100 & 3640 & 1965 \\
& LesionSegL & 6631 & 1843 & 734  & 200 & 7365 & 2043 \\
& LesionSegP & 6888 & 2296 & 2277 & 759 & 9165 & 3055 \\
\midrule
Chest
& LesionSegLung & 7755 & 4393 & 395 & 100 & 8150 & 4493 \\
\bottomrule
\end{tabular}
}
\end{table*}

\subsubsection{Experimental settings}
We consider three experimental settings, each corresponding to a SuG model trained using the same progressive three-stage pipeline. The three settings differ only in the pairing between the specialist and generalist training datasets. Specifically, \textbf{Setting 1} corresponds to the abdominal CT scenario using LesionSegAbdomen together with MedVL-CT69K; \textbf{Setting 2} also targets the abdominal CT scenario, but pairs LesionSegAbdomen with Merlin; and \textbf{Setting 3} corresponds to the chest CT scenario using LesionSegLung together with CT-RATE. In each setting, the lesion segmentation dataset provides voxel-level specialist supervision for learning anatomy- and lesion-aware representations, while the image–report dataset supports generalist vision–language learning. This design allows us to evaluate SuG under both abdominal and chest scenarios, as well as across different image–report sources.

\subsection{Baselines and implementation details}
\label{app_sec:impl_detail}
\subsubsection{Baselines}
We compare SuG against three categories of methods: specialist segmentation models, generalist vision–language models, and specialist–generalist synergy (S+G) methods that integrate segmentation-based specialist modeling with vision–language learning.
For the specialist baseline, we adopt nnU-Net~\cite{nnUNet}, a widely used supervised segmentation framework, and train it for anatomy-specific lesion-centric diagnosis. The procedure used to derive diagnostic probabilities from its segmentation outputs is described in Appendix~\ref{app_sec:seg_to_cls_score}.
The generalist baselines include CLIP~\cite{CLIP}, LOVT~\cite{LOVT}, Imitate~\cite{imitate}, BIUD~\cite{BIUD}, Merlin~\cite{Merlin}, fVLM~\cite{fvlm}, and ViSD-Boost~\cite{VISD-Boost}.
As representative specialist–generalist synergy baselines, we include three methods: OpenVocabCT~\cite{OpenVocabCT}, HCFNet~\cite{HCFNet}, and VLWS~\cite{VLWS}. It is worth noting that existing specialist–generalist synergy methods are mainly developed to enhance specialist, rather than to strengthen the broad diagnostic capability of generalist models. We therefore regard these methods as the closest available baselines for specialist–generalist collaboration, although they are not specifically designed for the super-generalist setting considered in this work. This gap also highlights the value of super-generalist that explicitly unifies broad diagnostic coverage with specialist-level lesion understanding.
All baseline methods are re-trained on the same training data used by SuG for fair comparison.

\subsubsection{Implementation details}
\textbf{Stage 1: Specialist pretraining}. In the first stage, we train the specialist pathway within the nnU-Net framework. The architecture is automatically configured by nnU-Net. To maintain consistency with the subsequent vision–language alignment stage, we use a target spacing of $[5.0, 1.0, 1.0]$ and a patch size of $96 \times 256 \times 384$~\cite{fvlm, VISD-Boost}. Training is performed for 1000 epochs using SGD with a batch size of 2. The learning rate is initialized at $1 \times 10^{-2}$ and decayed to 0 over the course of training. This stage is trained on a single NVIDIA H20 GPU.
\textbf{Stage 2: Specialist-enhanced vision–language alignment}. In the second stage, we initialize the visual encoder from the pretrained specialist model obtained in Stage 1 and perform specialist-enhanced vision–language learning. A pretrained BERT~\cite{devlin2019bert} is adopted as the text encoder. We jointly optimize the segmentation objective and the vision–language alignment objective for 30 epochs using AdamW with a batch size of 48. The learning rate is initialized at $1 \times 10^{-4}$ and decayed to $1 \times 10^{-6}$ with cosine annealing. This stage is trained on 24 NVIDIA H20 GPUs.
\textbf{Stage 3: Lesion-guided attention calibration}. In the final stage, we initialize the model from Stage 2, introduce the lesion-guided attention calibration loss, and continue training for 5 epochs. Unless otherwise specified, all remaining training settings are identical to those used in Stage 2. We implement all models in Python 3.11 using PyTorch 2.6.0 and CUDA 12.4.

\subsection{Diagnostic score derivation for specialist baselines}
\label{app_sec:seg_to_cls_score}

Our specialist baselines are built on nnU-Net segmentation models, which produce voxel-wise lesion predictions rather than scan-level disease scores. To evaluate these models under classification metrics, we convert their segmentation outputs into diagnostic scores.
For an input CT scan, the segmentation model outputs both a segmentation map and the corresponding class probability map. We first perform connected-component analysis on the segmentation map to identify all spatially independent lesion regions for the target class. For each connected lesion region, the lesion score is defined as the average probability of the target class over all voxels within that region.
If a CT scan contains multiple disconnected lesions of the same class, we use the maximum lesion score among all such regions as the final diagnostic score for that disease in the scan. This strategy allows voxel-wise specialist models to be directly compared with other methods under classification metrics.

\subsection{Evaluation tasks and metrics}
\subsubsection{Evaluation tasks}
We design five complementary evaluation tasks to assess whether SuG satisfies the three key requirements of a super-generalist model: broad diagnostic coverage, specialist-competitive lesion-centric diagnosis, and reliable lesion grounding.

\textbf{General disease diagnosis}. We first evaluate the broad-coverage diagnostic capability of SuG on multi-disease test sets to examine whether it can serve as a strong generalist model. This evaluation is conducted on MedVL-CT69K, Merlin, CT-RATE, and RAD-ChestCT, covering both abdominal and chest CT scenarios.

\textbf{Fine-grained lesion-centric diagnosis}. We further evaluate SuG on lesion-centric disease categories that require fine-grained lesion perception and are traditionally challenging for generalist models. We include four lesion-centric categories from LesionSegAbdomen, as well as four lesion-related disease categories from the 54 test diseases in MedVL-CT69K. Detailed label mappings are provided in the Table~\ref{app_tab:four_cancer_mapping}.

\textbf{Annotated lesion grounding}. To evaluate whether SuG can localize lesions that receive supervision during training, we assess lesion grounding primarily in the abdominal setting. Specifically, we evaluate grounding performance on four annotated lesion categories using the LesionSegAbdomen test set.

\textbf{Generalization to unannotated lesions}. Beyond the supervised lesion categories, we further evaluate whether SuG can generalize lesion grounding to unannotated lesions. This evaluation is conducted on MedVL-CT69K, where lesion categories without direct specialist supervision are used to assess generalization.

\textbf{Radiology report generation}. Finally, we evaluate whether the specialist-enhanced visual representations learned by SuG can transfer to cross-modal medical image interpretation beyond diagnosis and grounding. To do so, we pair the pretrained vision encoder with a text decoder and assess its performance on the downstream radiology report generation task.

\subsubsection{Evaluation metrics}
\textbf{Disease diagnosis tasks:} Following previous works~\cite{fvlm, Merlin,  CT-RATE, VISD-Boost}, we report AUC, F1-score, sensitivity (SE), and specificity (SP).
\textbf{Annotated and unannotated lesion grounding:} We perform qualitative and quantitative evaluations. Qualitatively, we visualize image–text similarity maps (attention maps) to examine whether the model attends to lesion-relevant regions. Quantitatively, we compare the predicted heatmaps with lesion masks using three voxel-level metrics. Specifically, AUC measures overall discriminability between lesion and background voxels across all thresholds; AUPR is particularly informative under class imbalance and reflects the precision–recall trade-off for lesion localization; and Cohen’s d quantifies the standardized separation between heatmap responses on lesion versus non-lesion voxels. Higher values indicate more accurate lesion localization and clearer separation between lesion and background regions.
\textbf{Radiology report generation:} We report both clinical efficacy metrics and conventional NLG metrics. The former include precision, recall, F1-score, and GREEN, while the latter include BLEU4, ROUGE-L, METEOR, and CIDEr.

\begin{table}[t]
\centering
\caption{\textbf{Label mapping from MedVL-CT69K to the four tumor categories used for evaluation.}}
\label{app_tab:four_cancer_mapping}
\begin{tabular}{p{3.0cm}|p{4.8cm}}
\toprule
Four-tumor category & Original labels \\
\midrule
Large bowel tumor & rectal cancer, colon cancer \\
Liver tumor & liver cyst, liver cancer, liver metastatic tumor \\
Stomach tumor & gastric cancer \\
Pancreas tumor & pancreatic cancer \\
\bottomrule
\end{tabular}
\end{table}

\begin{table*}[t]
\centering
\caption{\textbf{Disease subsets defined for validation-set evaluation.} The "General diseases" set comprises all 36 disease labels. The "Unannotated localized lesions" subset consists of diseases not explicitly covered by the four-tumor specialist supervision, while the "Annotated lesions" subset corresponds to the four tumor categories used for lesion-centric evaluation.}
\label{app_tab:val_disease_lists}
\renewcommand{\arraystretch}{1.15}
\setlength{\tabcolsep}{4pt}
\resizebox{\textwidth}{!}{
\begin{tabular}{p{3.2cm}|p{13.8cm}}
\toprule
Subset & Disease labels \\
\midrule
General diseases &
liver\_cyst, liver\_fatty liver, liver\_cirrhosis, liver\_intrahepatic bile duct dilatation, liver\_metastatic tumor, liver\_Glisson's sheath fluid, liver\_liver cancer, portal vein\_embolism, portal vein\_portal hypertension, kidney\_cyst, kidney\_stone, kidney\_hydronephrosis, kidney\_atrophy, adrenal gland\_thickening, adrenal gland\_nodule, spleen\_splenomegaly, spleen\_infarction, spleen\_hemangioma, gallbladder\_adenomyomatosis, gallbladder\_stone, gallbladder\_inflammation, stomach\_gastric cancer, stomach\_wall thickening, pancreas\_atrophy, pancreas\_pancreatic duct dilatation, pancreas\_inflammation, pancreas\_pancreatic cancer, large bowel\_appendicitis, large bowel\_rectal cancer, large bowel\_obstruction, large bowel\_colon cancer, large bowel\_fluid accumulation, large bowel\_gas accumulation, small bowel\_obstruction, small bowel\_gas accumulation, small bowel\_fluid accumulation \\
\midrule
Unannotated localized lesions &
liver\_Glisson's sheath fluid, portal vein\_embolism, kidney\_cyst, kidney\_stone, adrenal gland\_nodule, spleen\_infarction, spleen\_hemangioma, gallbladder\_adenomyomatosis, gallbladder\_stone, large bowel\_appendicitis \\
\midrule
Annotated lesions &
large bowel\_tumor, liver\_tumor, stomach\_tumor, pancreas\_tumor \\
\bottomrule
\end{tabular}
}
\end{table*}

\begin{table}[t]
\centering
\small
\setlength{\tabcolsep}{10pt}
\renewcommand{\arraystretch}{1.05}
\begin{tabular}{ll}
\toprule
Abbreviation & Full name \\
\midrule
Adr & Adrenal gland \\
Bla & Bladder \\
Col & Colon \\
Eso & Esophagus \\
GB  & Gallbladder \\
Hrt & Heart \\
Kid & Kidney \\
Liv & Liver \\
Lng & Lung \\
Pan & Pancreas \\
PV  & Portal vein \\
SI  & Small intestine \\
Spl & Spleen \\
Sto & Stomach \\
Sac & Sacrum \\
Avg & Average \\
\bottomrule
\end{tabular}
\caption{\textbf{Full names of anatomical abbreviations used in Table~\ref{tab1}.}}
\label{tab:anatomy_abbrev}
\end{table}

\section{Merlin Training and Inference Protocol}
\label{app:merlin_protocol}

Merlin contains both anatomy-grounded disease labels that naturally map to one of our anatomical targets and scan-level labels that are not naturally associated with a single anatomy. To handle both cases in a unified framework, we use a hybrid training and inference protocol.

\subsection{Alternating training objectives on Merlin}
\label{app:merlin_training}

In addition to the anatomy-wise contrastive learning described in the main paper, Merlin training alternates between two alignment modes:

\paragraph{(1) Anatomy-wise alignment.}
For anatomy-grounded diseases, we use the same anatomy-wise contrastive strategy as in the main paper: anatomy-specific visual embeddings extracted from segmented anatomy regions are aligned with anatomy-level textual descriptions.

\paragraph{(2) Whole-body image--report alignment.}
For whole-body diseases, we additionally perform whole-body image--report contrastive learning. In this branch, the full CT volume is encoded into a global visual representation and aligned with the complete report text. The same bidirectional image--text contrastive objective is used, but applied at scan level rather than anatomy level.
In practice, these two alignment modes are alternated during Merlin training. This enables the model to learn both anatomy-specific and scan-level semantics.

\subsection{Inference on Merlin}
\label{app:merlin_inference}

At test time, we select the inference mode according to the disease subset (see Table~\ref{app_tab:merlin_label_partition}):\\
\textbf{Anatomy-level inference.}
For anatomy-wise diseases, we extract the visual embedding of the relevant anatomy and compute similarity to the corresponding disease prompt embedding.\\
\textbf{Whole-body inference.}
For whole-body diseases, we compute a global visual embedding from the full CT and compare it with the disease prompt embedding.

\begin{table*}[t]
\centering
\caption{\textbf{Partition of Merlin diseases for the hybrid training and inference protocol.} Diseases are categorized into "Anatomy-wise" and "Whole-body" subsets, which are evaluated using anatomy-level and whole-body inference, respectively.}
\label{app_tab:merlin_label_partition}
\footnotesize
\renewcommand{\arraystretch}{1.10}
\setlength{\tabcolsep}{5pt}

\begin{tabularx}{\textwidth}{@{}p{2.6cm}X@{}}
\toprule
Subset & Disease list \\
\midrule
Anatomy-wise &
{\ttfamily
aortic\_valve\_calcification, pleural\_effusion, splenomegaly, abdominal\_aortic\_aneurysm, hepatic\_steatosis, bowel\_obstruction, hepatomegaly, cardiomegaly, pancreatic\_atrophy, hydronephrosis, surgically\_absent\_gallbladder, atherosclerosis, submucosal\_edema, hiatal\_hernia, atelectasis, biliary\_ductal\_dilation, renal\_cyst, renal\_hypodensities, gallstones, fracture, appendicitis
} \\
\midrule
Whole-body &
{\ttfamily
free\_air, metastatic\_disease, lymphadenopathy, anasarca, ascites, coronary\_calcification, thrombosis, prostatomegaly, osteopenia
} \\
\bottomrule
\end{tabularx}
\end{table*}

\begin{figure*}
    \centering
    \includegraphics[width=0.8\linewidth]{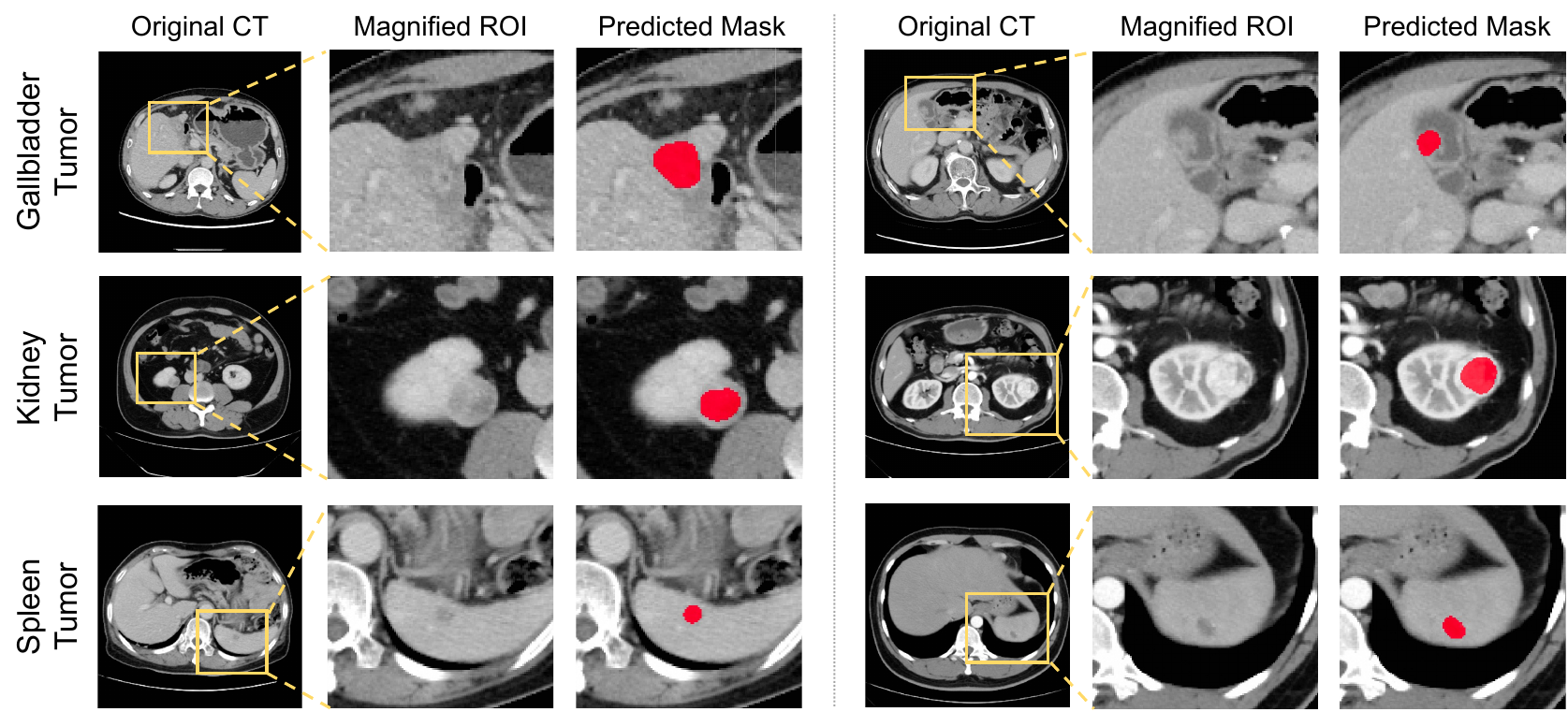}
    \caption{\textbf{Visualization of unannotated lesion segmentation.} For each representative case, the original CT slice, corresponding magnified ROI, and predicted lesion mask are presented.}
    \label{app_fig:pan_seg}
\end{figure*}

\section{Radiology report generation transfer}
\label{app:report_generation}

To further assess whether the diagnosis-aware and lesion-sensitive visual representations learned by SuG can transfer beyond classification and grounding, we evaluate SuG on a downstream radiology report generation task. Specifically, we pair the pretrained vision encoder with a text decoder and assess the resulting model under two transfer settings: \emph{Frozen}, where the visual encoder is fixed, and \emph{Finetuning}, where the visual encoder is jointly optimized with the text decoder. For comparison, we include visual pretraining baselines (Supervised and MAE), generalist vision–language baselines (CLIP, BIUD, Merlin, fVLM, and ViSD-Boost), and specialist–generalist synergy baselines (OpenVocabCT, HCFNet, and VLWS).

Table~\ref{tab:report_generation} shows that SuG achieves strong performance across both clinical efficacy metrics and natural language generation metrics. Under the Frozen setting, SuG obtains the best F1-score of 36.1, demonstrating that its pretrained visual encoder can effectively captures highly transferable clinical information. Under the Finetuning setting, SuG further improves the clinical quality of the generated reports, raising the F1-score from 36.1 to 42.2, while remaining competitive on the other evaluation metrics. Overall, these results indicate that SuG could provides clinically informative visual representations, enabling faithful downstream report generation.

\begin{table*}[ht!]
\centering
\centering
\tiny
\resizebox{0.89\linewidth}{!}{
\setlength{\tabcolsep}{1.5mm}{
\begin{tabular}{c|cccccccccc}
\toprule
Encoder  & Init                      & Persion             & Recal             & F1            & GREEN         & BLEU4         & ROUGE-L       & METEOR        & CIDEr          \\ \midrule
\multirow{9}{*}{Frozen} &
 Supervised                         & 19.1          & 18.6          & 13.2          & 25.9          & 12.8          & 40.6          & 30.8          & 6.6            \\
 & MAE~\cite{mae}                   & 8.9           & 5.9           & 4.3           & 21.6          & 13.1          & 41.6          & 30.5          & 6.1            \\
 & CLIP~\cite{CLIP}                 & 21.6          & 20.4          & 14.6          & 33.4          & 15.5          & 42.2          & 31.0          & 9.6            \\
 & BIUD~\cite{BIUD}                 & 17.0          & 21.4          & 15.9          & 33.7          & 18.9          & 44.2          & 29.1          & 13.9           \\
 & Merlin~\cite{Merlin}             & 22.6          & 20.9          & 20.7          & 34.2          & 19.0          & 43.8          & 30.0          & 14.3           \\
 & fVLM~\cite{fvlm} & 24.0          & 31.6          & 26.5          & 37.2          & 19.6          & 45.1          & 31.3          & 14.9           \\
 & ViSD-Boost~\cite{VISD-Boost}     & 34.3          & \textbf{39.3}          & \underline{35.2}          & 44.4          & \underline{24.7}          & \underline{48.7}          & \textbf{32.7}          & \underline{27.3}           \\ 
 & OpenVocabCT~\cite{OpenVocabCT}   & 36.2 & 37.7 & 34.9 & \underline{44.9} & 23.9 & 48.1 & 32.2 & 23.0  \\
 & HCFNet~\cite{HCFNet}             & 36.3 & 36.8 & 34.3 & 44.6 & 24.3 & 48.3 & 32.2 & 23.9  \\
 & VLWS~\cite{VLWS}                 & \textbf{36.8} & 37.2 & 35.1 & \textbf{45.3} & 24.5 & 48.6 & 32.3 & 23.1  \\
 & \textbf{SuG (Ours)}                              & \underline{36.7}          & \underline{38.2}          & \textbf{36.1}          & 44.5          & \textbf{25.1}          & \textbf{48.8}          & \underline{32.6}          & \textbf{28.0}           \\  \midrule
\multirow{9}{*}{Finetuning} &
 Supervised                         & 18.0          & 28.3          & 20.4          & 35.5          & 17.9          & 43.4          & 30.6          & 11.7           \\
 & MAE~\cite{mae}                   & 13.4          & 14.1          & 10.9          & 29.4          & 15.1          & 42.5          & 30.3          & 8.8            \\
 & CLIP~\cite{CLIP}                 & 21.0          & 29.5          & 23.2          & 37.6          & 19.5          & 44.8          & 30.7          & 14.3           \\
 & BIUD~\cite{BIUD}                 & 26.1          & 31.6          & 24.2          & 38.8          & 19.0          & 44.7          & 30.9          & 13.9           \\
 & Merlin~\cite{Merlin}             & 27.5          & 29.9          & 25.8          & 39.2          & 20.9          & 46.0          & 31.1          & 17.2           \\
 & fVLM~\cite{fvlm} & 38.6          & 36.9          & 32.7          & 40.2          & 21.9          & 46.4          & 31.6          & 17.1           \\
 & ViSD-Boost~\cite{VISD-Boost}     & \underline{39.8}          & 44.1          & 40.9          & 46.7          & \textbf{28.4}          & \textbf{51.0}          & \underline{34.1}          & \textbf{50.7}           \\ 
 & OpenVocabCT~\cite{OpenVocabCT}   & 38.2 & 44.2 & 40.4 & 47.5 & 26.2 & 49.7 & 33.6 & 34.5  \\
 & HCFNet~\cite{HCFNet}             & \textbf{39.9} & 43.7 & 40.8 & \textbf{49.1} & \underline{27.5} & 50.5 & 34.0 & 36.4  \\
 & VLWS~\cite{VLWS}                 & \underline{39.8} & \underline{44.4} & \underline{41.1} & 47.4 & 27.2 & 50.3 & 34.0 & 37.7  \\
 & \textbf{SuG (Ours)}                              & 39.6          & \textbf{46.6}          & \textbf{42.2}          & \underline{48.7}          & \underline{27.5}          & \underline{50.6}          & \textbf{34.2}          & \underline{47.8}           \\  \bottomrule
\end{tabular}}}
\caption{\textbf{Radiology report generation performance on the MedVL-CT69K test set.} We compare two transfer settings: \textit{Frozen} (fixed visual encoder) and \textit{Finetuning} (jointly optimized visual encoder and report decoder). "Supervised" refers to a classification model trained on 54 diseases, while "MAE" and "Supervised" represent 3D models pre-trained on the MedVL-CT69K training set. \textbf{Bold} and \underline{underlined} values denote the best and second-best scores, respectively.}
\label{tab:report_generation}
\end{table*}

\section{Statistical Significance Analysis}
\label{sec:appendix_significance}

To further validate the robustness of the performance gains in the main results, we conduct statistical significance analysis using paired bootstrap resampling. For each benchmark, we compare SuG against the \emph{second-best} competing method under the corresponding evaluation setting. Specifically, the compared methods are HCFNet on CT-RATE, RAD-ChestCT, and Merlin, and VLWS on MedVL-CT69K and LesionSegAbdomen.

Our significance analysis is performed at the case level or patient level, following the protocol used in each dataset. For each bootstrap replicate, we resample evaluation cases with replacement in a paired manner for the two compared methods, recompute the per-dataset macro-level metrics, and record the performance difference between SuG and the second-best method. We then report the two-sided bootstrap $p$-value.

Tables~\ref{tab:medvl_bootstrap}--\ref{tab:lesionsegabdomen_bootstrap} summarize the statistical significance results on all evaluated benchmarks. Overall, the improvements of SuG over the second-best method are statistically significant on the vast majority of metrics. In particular, strong significance is consistently observed for AUC-related performance across datasets, and most of the corresponding sensitivity, specificity, F1, and precision gains are also significant. Only a small number of metrics, such as Precision on CT-RATE and ACC on RAD-ChestCT, do not reach statistical significance.

\begin{table}[htbp]
\centering
\caption{\textbf{Bootstrap significance analysis on MedVL-CT69K.} 
We use patient-level paired bootstrap resampling to evaluate the statistical significance of macro-level performance differences between SuG and VLWS. Significance levels are denoted as follows: $* p < 0.05, ** p < 0.01, *** p < 0.001$.}
\resizebox{0.56\linewidth}{!}{
\begin{tabular}{lccc}
\toprule
Metric & VLWS & SuG & p-value \\
\midrule
AUC         & 86.7 & 90.1 \textcolor{gray}{\footnotesize(+3.36)} & *** \\
Sensitivity & 81.8 & 85.0 \textcolor{gray}{\footnotesize(+3.20)} & *** \\
Specificity & 81.9 & 84.9 \textcolor{gray}{\footnotesize(+3.00)} & *** \\
\bottomrule
\end{tabular}
}
\label{tab:medvl_bootstrap}
\end{table}

\begin{table}[htbp]
\centering
\caption{\textbf{Bootstrap significance analysis on Merlin.} 
We use patient-level paired bootstrap resampling to evaluate the statistical significance of macro-level performance differences between SuG and HCFNet for the balanced metrics . Significance levels are denoted as follows: $* p < 0.05, ** p < 0.01, *** p < 0.001$.}
\resizebox{0.56\linewidth}{!}{
\begin{tabular}{lccc}
\toprule
Metric & HCFNet & SuG & p-value \\
\midrule
Balanced AUC & 83.2 & 85.2 \textcolor{gray}{\footnotesize(+2.02)} & ** \\
Balanced F1  & 78.7 & 80.8 \textcolor{gray}{\footnotesize(+2.05)} & ** \\
\bottomrule
\end{tabular}
}
\label{tab:merlin_bootstrap}
\end{table}

\begin{table}[htbp]
\centering
\caption{\textbf{Bootstrap significance analysis on Merlin for unbalanced metrics.} 
We use patient-level paired bootstrap resampling to evaluate the statistical significance of macro-level performance differences between SuG and HCFNet for the unbalanced metrics. Significance levels are denoted as follows: $* p < 0.05, ** p < 0.01, *** p < 0.001$}
\resizebox{0.56\linewidth}{!}{
\begin{tabular}{lccc}
\toprule
Metric & HCFNet & SuG & p-value \\
\midrule
AUC & 82.9 & 85.1 \textcolor{gray}{\footnotesize(+2.19)} & *** \\
F1  & 57.2 & 60.7 \textcolor{gray}{\footnotesize(+3.44)} & *** \\
\bottomrule
\end{tabular}
}
\label{tab:merlin_unbalanced_bootstrap}
\end{table}

\begin{table}[htbp]
\centering
\caption{\textbf{Bootstrap significance analysis on CT-RATE.} 
We use patient-level paired bootstrap resampling to evaluate the statistical significance of macro-level performance differences between SuG and HCFNet. Significance levels are denoted as follows: $* p < 0.05, ** p < 0.01, *** p < 0.001$.}
\resizebox{0.56\linewidth}{!}{
\begin{tabular}{lccc}
\toprule
Metric & HCFNet & SuG & p-value \\
\midrule
AUC       & 80.3 & 81.1 \textcolor{gray}{\footnotesize(+0.80)} & *** \\
ACC       & 73.5 & 74.4 \textcolor{gray}{\footnotesize(+0.84)} & *** \\
F1        & 76.4 & 77.1 \textcolor{gray}{\footnotesize(+0.69)} & *** \\
Precision & 40.7 & 40.8 \textcolor{gray}{\footnotesize(+0.15)} &  \\
\bottomrule
\end{tabular}
}
\label{tab:ctrate_bootstrap}
\end{table}

\begin{table}[htbp]
\centering
\caption{\textbf{Bootstrap significance analysis on RAD-ChestCT.} 
We use case-level paired bootstrap resampling to evaluate the statistical significance of macro-level performance differences between SuG and HCFNet on RAD-ChestCT. Significance levels are denoted as follows: $* p < 0.05, ** p < 0.01, *** p < 0.001$.}
\resizebox{0.56\linewidth}{!}{
\begin{tabular}{lccc}
\toprule
Metric & HCFNet & SuG & p-value \\
\midrule
AUC       & 71.7 & 73.4 \textcolor{gray}{\footnotesize(+1.69)} & *** \\
ACC       & 67.8 & 68.1 \textcolor{gray}{\footnotesize(+0.31)} &  \\
F1        & 71.5 & 71.9 \textcolor{gray}{\footnotesize(+0.36)} & * \\
Precision & 35.9 & 36.9 \textcolor{gray}{\footnotesize(+0.99)} & *** \\
\bottomrule
\end{tabular}
}
\label{tab:radchest_bootstrap}
\end{table}

\begin{table}[htbp]
\centering
\caption{\textbf{Bootstrap significance analysis on MedVL-CT69K for the four tumor categories.} 
We use patient-level paired bootstrap resampling across the four tumor categories to evaluate the statistical significance of macro-level performance differences between SuG and VLWS. Significance levels are denoted as follows: $* p < 0.05, ** p < 0.01, *** p < 0.001$.}
\resizebox{0.56\linewidth}{!}{
\begin{tabular}{lccc}
\toprule
Metric & VLWS & SuG & p-value \\
\midrule
AUC         & 88.1 & 93.7 \textcolor{gray}{\footnotesize(+5.63)} & *** \\
Sensitivity & 81.4 & 86.8 \textcolor{gray}{\footnotesize(+5.34)} & * \\
Specificity & 82.3 & 88.0 \textcolor{gray}{\footnotesize(+5.72)} & *** \\
\bottomrule
\end{tabular}
}
\label{tab:medvl_tumor_bootstrap}
\end{table}

\begin{table}[htbp]
\centering
\caption{\textbf{Bootstrap significance analysis on LesionSegAbdomen.} 
We use patient-level paired bootstrap resampling across the four tumor categories to evaluate the statistical significance of macro-level performance differences between SuG and VLWS. Significance levels are denoted as follows: $* p < 0.05, ** p < 0.01, *** p < 0.001$.}
\resizebox{0.56\linewidth}{!}{
\begin{tabular}{lccc}
\toprule
Metric & VLWS & SuG & p-value \\
\midrule
AUC         & 87.0 & 96.1 \textcolor{gray}{\footnotesize(+9.10)} & *** \\
Sensitivity & 78.9 & 89.4 \textcolor{gray}{\footnotesize(+10.50)} & *** \\
Specificity & 81.9 & 91.7 \textcolor{gray}{\footnotesize(+9.80)} & *** \\
\bottomrule
\end{tabular}
}
\label{tab:lesionsegabdomen_bootstrap}
\end{table}

\begin{table*}[!htbp]
\centering
\footnotesize
\setlength{\tabcolsep}{2.2pt}
\renewcommand{\arraystretch}{1.12}

\resizebox{\textwidth}{!}{%
\begin{tabular}{
l c
*{8}{S[table-format=2.1] S[table-format=2.1] S[table-format=2.1]}
}
\toprule
\multirow{2}{*}{Method} & \multirow{2}{*}{Type} &
\multicolumn{3}{c}{Adrenal gland} &
\multicolumn{3}{c}{Bladder} &
\multicolumn{3}{c}{Colon} &
\multicolumn{3}{c}{Esophagus} &
\multicolumn{3}{c}{Gallbladder} &
\multicolumn{3}{c}{Heart} &
\multicolumn{3}{c}{Kidney} &
\multicolumn{3}{c}{Liver} \\
\cmidrule(lr){3-5}\cmidrule(lr){6-8}\cmidrule(lr){9-11}\cmidrule(lr){12-14}%
\cmidrule(lr){15-17}\cmidrule(lr){18-20}\cmidrule(lr){21-23}\cmidrule(lr){24-26}
& & {SE} & {SP} & {AUC} & {SE} & {SP} & {AUC} & {SE} & {SP} & {AUC} &
    {SE} & {SP} & {AUC} & {SE} & {SP} & {AUC} & {SE} & {SP} & {AUC} &
    {SE} & {SP} & {AUC} & {SE} & {SP} & {AUC} \\
\midrule
CLIP~\cite{CLIP}                            & G   & 66.6 & 55.4 & 63.2 & 57.7 & 67.6 & 65.1 & 64.4 & 63.2 & 65.8 & 65.1 & 78.3 & 67.3 & 55.5 & 59.9 & 59.5 & 36.2 & 77.1 & 44.1 & 55.5 & 61.6 & 59.9 & 70.8 & 65.4 & 72.4 \\
LOVT~\cite{LOVT}                            & G   & 61.8 & 54.6 & 60.6 & 71.4 & 62.3 & 70.9 & 69.6 & 58.5 & 67.5 & 84.2 & 85.1 & 89.3 & 65.2 & 51.8 & 61.2 & 84.6 & 64.9 & 78.3 & 62.2 & 54.7 & 60.2 & 68.3 & 60.2 & 69.3 \\
Imitate~\cite{imitate}                      & G   & 64.3 & 55.9 & 60.2 & 72.6 & 67.9 & 74.1 & 69.2 & 60.7 & 68.0 & 98.1 & 89.4 & 95.6 & 60.0 & 59.7 & 62.5 & 71.5 & 75.1 & 70.7 & 60.7 & 55.8 & 59.9 & 66.5 & 65.4 & 69.8 \\
BIUD~\cite{BIUD}                            & G   & 64.9 & 56.0 & 63.4 & 79.8 & 73.5 & 81.0 & 70.4 & 64.8 & 70.0 & 54.1 & 91.9 & 62.6 & 60.6 & 61.1 & 64.2 & 68.2 & 56.1 & 62.1 & 60.8 & 61.8 & 63.7 & 72.6 & 74.0 & 79.2 \\
Merlin~\cite{Merlin}                        & G   & 58.9 & 57.9 & 60.3 & 70.8 & 73.0 & 76.9 & 71.1 & 62.0 & 69.1 & 41.5 & 88.5 & 49.2 & 64.6 & 53.5 & 61.2 & 69.6 & 75.1 & 72.8 & 58.6 & 64.5 & 64.2 & 73.6 & 75.9 & 80.1 \\
fVLM~\cite{fvlm}                            & G   & 63.0 & 63.9 & 65.7 & 76.2 & 77.3 & 84.0 & 76.1 & 75.1 & 80.8 & 94.4 & 96.1 & 98.2 & 64.9 & 58.8 & 64.8 & 87.2 & 75.8 & 85.8 & 67.9 & 72.5 & 74.5 & 77.2 & 76.0 & 82.5 \\
ViSD-Boost~\cite{VISD-Boost}                & G   & 63.5 & 64.9 & 68.5 & 75.0 & 74.4 & 81.2 & 78.2 & 75.7 & 81.9 & 94.4 & 95.4 & 98.1 & 65.6 & 69.7 & 72.6 & 84.6 & 82.6 & 90.5 & 72.4 & 74.5 & 78.5 & 78.4 & 80.3 & 85.9 \\
\midrule
OpenVocabCT~\cite{OpenVocabCT}              & S+G & 60.2 & 67.6 & 68.8 & 81.0 & 74.4 & 84.2 & 77.0 & 81.1 & 84.3 & 88.7 & 98.3 & 96.3 & 77.6 & 78.6 & 85.0 & 75.0 & 81.1 & 81.5 & 77.3 & 79.2 & 84.0 & 84.5 & 83.3 & 90.6  \\
HCFNet~\cite{HCFNet}                        & S+G & 70.1 & 63.7 & 71.2 & 81.0 & 87.2 & 89.2 & 75.6 & 77.9 & 83.0 & 93.7 & 97.9 & 99.0 & 80.5 & 74.8 & 83.5 & 82.9 & 76.2 & 85.6 & 75.5 & 81.8 & 84.4 & 81.4 & 85.2 & 89.3  \\
VLWS~\cite{VLWS}                            & S+G & 68.2 & 64.5 & 70.1 & 81.0 & 82.8 & 88.9 & 79.6 & 79.3 & 84.1 & 93.7 & 94.7 & 95.3 & 80.9 & 78.5 & 85.6 & 86.6 & 80.6 & 89.5 & 71.8 & 79.1 & 80.4 & 79.4 & 79.9 & 84.9  \\
\midrule
\textbf{SuG (Ours)}	                                        & S+G & 71.7 & 82.6 & 82.9 & 89.3 & 84.5 & 92.1 & 80.8 & 81.6 & 87.0 & 92.4 & 93.3 & 97.2 & 81.9 & 75.9 & 84.9 & 87.8 & 84.4 & 91.6 & 80.2 & 83.0 & 87.4 & 87.4 & 86.9 & 92.2  \\
\midrule
\multirow{2}{*}{Method} & \multirow{2}{*}{Type} &
\multicolumn{3}{c}{Lung} &
\multicolumn{3}{c}{Pancreas} &
\multicolumn{3}{c}{Portal vein} &
\multicolumn{3}{c}{Small Intestine} &
\multicolumn{3}{c}{Spleen} &
\multicolumn{3}{c}{Stomach} &
\multicolumn{3}{c}{Sacrum} &
\multicolumn{3}{c}{Average} \\
\cmidrule(lr){3-5}\cmidrule(lr){6-8}\cmidrule(lr){9-11}\cmidrule(lr){12-14}%
\cmidrule(lr){15-17}\cmidrule(lr){18-20}\cmidrule(lr){21-23}\cmidrule(lr){24-26}
& & {SE} & {SP} & {AUC} & {SE} & {SP} & {AUC} & {SE} & {SP} & {AUC} &
    {SE} & {SP} & {AUC} & {SE} & {SP} & {AUC} & {SE} & {SP} & {AUC} &
    {SE} & {SP} & {AUC} & {SE} & {SP} & {AUC} \\
\midrule
CLIP~\cite{CLIP}                            & G   & 80.4 & 96.1 & 88.3 & 65.4 & 62.4 & 65.0 & 72.4 & 72.4 & 78.6 & 69.9 & 66.6 & 74.5 & 72.8 & 65.9 & 71.1 & 62.5 & 68.0 & 68.6 & 47.1 & 56.0 & 47.0 & 65.5 & 68.0 & 68.4 \\
LOVT~\cite{LOVT}                            & G   & 78.7 & 65.0 & 80.9 & 68.3 & 62.5 & 67.8 & 82.6 & 60.2 & 75.5 & 72.4 & 61.5 & 70.5 & 70.1 & 49.0 & 66.1 & 62.9 & 67.9 & 69.1 & 70.6 & 38.8 & 48.9 & 70.8 & 60.1 & 69.4 \\
Imitate~\cite{imitate}                      & G   & 81.1 & 89.7 & 86.7 & 65.0 & 61.3 & 64.3 & 76.1 & 69.3 & 80.5 & 76.1 & 68.0 & 77.6 & 64.0 & 68.9 & 71.3 & 64.0 & 63.7 & 66.3 & 35.3 & 43.4 & 29.0 & 69.2 & 66.6 & 70.6 \\
BIUD~\cite{BIUD}                            & G   & 69.3 & 84.5 & 72.1 & 72.4 & 70.3 & 76.9 & 82.5 & 71.4 & 82.2 & 74.8 & 67.5 & 75.1 & 65.6 & 72.3 & 72.3 & 63.1 & 63.8 & 66.1 & 70.6 & 29.2 & 43.8 & 69.3 & 69.0 & 71.4 \\
Merlin~\cite{Merlin}                        & G   & 76.9 & 80.1 & 78.7 & 74.2 & 63.8 & 73.5 & 86.2 & 78.0 & 85.9 & 73.4 & 72.1 & 78.4 & 67.3 & 72.2 & 72.0 & 63.3 & 67.7 & 69.9 & 47.1 & 65.8 & 48.2 & 69.2 & 69.7 & 71.9 \\
fVLM~\cite{fvlm}                            & G   & 74.3 & 78.9 & 82.2 & 75.8 & 80.8 & 85.3 & 90.8 & 93.2 & 96.7 & 74.0 & 78.6 & 82.1 & 76.5 & 78.0 & 82.0 & 69.9 & 67.8 & 74.1 & 88.2 & 83.3 & 87.5 & 75.8 & 76.5 & 81.3 \\
ViSD-Boost~\cite{VISD-Boost}                & G   & 90.2 & 88.9 & 92.7 & 80.7 & 85.1 & 88.9 & 92.7 & 92.7 & 97.3 & 84.1 & 79.1 & 88.3 & 78.2 & 77.1 & 82.9 & 73.1 & 77.4 & 81.1 & 70.6 & 75.7 & 77.5 & 79.4 & 79.7 & 84.9 \\
\midrule
OpenVocabCT~\cite{OpenVocabCT}              & S+G & 64.3 & 86.4 & 68.1 & 86.9 & 83.8 & 90.9 & 91.7 & 90.6 & 95.5 & 81.4 & 86.7 & 89.8 & 87.1 & 87.9 & 94.2 & 78.2 & 84.3 & 87.4 & 82.4 & 85.8 & 88.4 & 79.4 & 83.2 & 85.8  \\
HCFNet~\cite{HCFNet}                        & S+G & 84.9 & 87.6 & 89.3 & 83.0 & 87.9 & 90.2 & 93.6 & 93.7 & 97.4 & 85.8 & 80.1 & 88.3 & 86.6 & 85.8 & 92.6 & 74.2 & 69.9 & 76.6 & 76.5 & 49.2 & 58.7 & 81.4 & 81.8 & 86.7  \\
VLWS~\cite{VLWS}                            & S+G & 82.3 & 88.4 & 87.0 & 86.0 & 85.0 & 90.8 & 94.5 & 91.7 & 97.6 & 84.8 & 82.4 & 89.8 & 87.2 & 87.4 & 94.1 & 77.1 & 80.7 & 85.2 & 88.2 & 65.7 & 80.0 & 81.8 & 81.9 & 86.7  \\
\midrule
\textbf{SuG (Ours)}	                        & S+G & 88.1 & 86.5 & 90.9 & 84.3 & 85.7 & 90.4 & 94.5 & 93.9 & 98.4 & 83.9 & 84.4 & 88.9 & 91.6 & 90.5 & 95.6 & 80.5 & 83.4 & 88.4 & 88.2 & 80.3 & 88.2 & \textbf{85.0} & \textbf{84.9} & \textbf{90.1}  \\
\bottomrule
\end{tabular}%
}
\caption{\textbf{General disease diagnosis performance on the MedVL-CT69K test set.}
Results are reported for 15 individual anatomical structures, with the final column ("Average") representing the aggregated performance across all 54 diseases. SE, SP, and AUC denote sensitivity, specificity, and area under the receiver operating characteristic curve, respectively. ``G'', ``S'', and ``S+G'' represent generalist-only, specialist-only, and specialist-generalist synergy methods, respectively.}
\label{app_tab:medvl_generalist}
\end{table*}

\begin{table*}[t]
\centering
\footnotesize
\setlength{\tabcolsep}{2.2pt}
\renewcommand{\arraystretch}{1.12}

\resizebox{1.0\textwidth}{\height}{%
\begin{tabular}{
l c
*{5}{S[table-format=2.1] S[table-format=2.1] S[table-format=2.1]}
}
\toprule
\multicolumn{17}{c}{\textbf{MedVL-CT69K}}\\
\midrule
\multirow{2}{*}{Method} & \multirow{2}{*}{Type} &
\multicolumn{3}{c}{Colon} &
\multicolumn{3}{c}{Liver} &
\multicolumn{3}{c}{Pancreas} &
\multicolumn{3}{c}{Stomach} &
\multicolumn{3}{c}{Average} \\
\cmidrule(lr){3-5}\cmidrule(lr){6-8}\cmidrule(lr){9-11}\cmidrule(lr){12-14}\cmidrule(lr){15-17}
& & {SE} & {SP} & {AUC} & {SE} & {SP} & {AUC} & {SE} & {SP} & {AUC} &
    {SE} & {SP} & {AUC} & {SE} & {SP} & {AUC} \\
\midrule
nnUNet-c~\cite{nnUNet}              & S   & 80.1 & 80.8 & 87.2 & {--} & {--} & {--} & {--} & {--} & {--} & {--} & {--} & {--} & {--} & {--} & {--} \\
nnUNet-l~\cite{nnUNet}              & S   & {--} & {--} & {--} & 70.1 & 77.5 & 79.5 & {--} & {--} & {--} & {--} & {--} & {--} & {--} & {--} & {--} \\
nnUNet-p~\cite{nnUNet}              & S   & {--} & {--} & {--} & {--} & {--} & {--} & 75.9 & 72.5 & 78.6 & {--} & {--} & {--} & {--} & {--} & {--} \\
nnUNet-s~\cite{nnUNet}              & S   & {--} & {--} & {--} & {--} & {--} & {--} & {--} & {--} & {--} & 71.8 & 64.3 & 74.7 & {--} & {--} & {--} \\
nnUNet-fuse~\cite{nnUNet}           & S   & 80.1 & 80.8 & 87.2 & 70.1 & 77.5 & 79.5 & 75.9 & 72.5 & 78.6 & 71.8 & 64.3 & 74.7 & 74.5 & 73.8 & 80.0 \\
\midrule
CLIP~\cite{CLIP}                    & G   & 59.6 & 65.4 & 65.9 & 55.0 & 61.5 & 60.0 & 62.1 & 76.5 & 73.0 & 68.4 & 63.4 & 71.6 & 61.3 & 66.7 & 67.6 \\
LOVT~\cite{LOVT}                    & G   & 92.8 & 21.4 & 68.1 & 64.5 & 49.0 & 58.4 & 72.4 & 67.6 & 74.9 & 89.7 & 32.2 & 69.4 & 79.9 & 42.5 & 67.7 \\
Imitate~\cite{imitate}              & G   & 83.1 & 34.1 & 63.5 & 59.8 & 55.6 & 60.3 & 69.0 & 63.6 & 71.4 & 59.8 & 72.5 & 71.9 & 67.9 & 56.4 & 66.8 \\
BIUD~\cite{BIUD}                    & G   & 83.1 & 34.5 & 66.3 & 57.4 & 56.5 & 59.6 & 75.9 & 72.4 & 77.7 & 65.0 & 73.1 & 73.1 & 70.3 & 59.1 & 69.2 \\
Merlin~\cite{Merlin}                & G   & 68.1 & 51.2 & 62.6 & 65.8 & 54.6 & 63.2 & 75.9 & 65.0 & 74.2 & 69.2 & 59.8 & 66.5 & 69.7 & 57.6 & 66.6 \\
fVLM~\cite{fvlm}                    & G   & 66.3 & 69.7 & 73.6 & 62.3 & 56.1 & 62.2 & 82.8 & 76.4 & 85.8 & 70.9 & 72.5 & 75.7 & 70.6 & 68.7 & 74.3 \\
ViSD-Boost~\cite{VISD-Boost}        & G   & 74.1 & 78.0 & 83.8 & 62.4 & 58.1 & 64.5 & 93.1 & 82.6 & 94.7 & 72.6 & 82.9 & 84.2 & 75.6 & 75.4 & 81.8 \\
\midrule
OpenVocabCT~\cite{OpenVocabCT}      & S+G & 81.3 & 75.3 & 87.0 & 77.9 & 78.1 & 84.9 & 93.1 & 92.6 & 97.4 & 81.2 & 90.9 & 89.6 & 83.4 & 84.2 & 89.7  \\
HCFNet~\cite{HCFNet}                & S+G & 77.1 & 78.0 & 86.0 & 67.3 & 73.1 & 75.1 & 89.7 & 94.2 & 94.8 & 84.6 & 80.4 & 89.3 & 79.7 & 81.4 & 86.3  \\
VLWS~\cite{VLWS}                    & S+G & 72.3 & 80.3 & 83.3 & 74.7 & 80.2 & 84.0 & 93.1 & 88.8 & 96.4 & 85.5 & 79.7 & 88.6 & 81.4 & 82.3 & 88.1  \\
\midrule
\textbf{SuG (Ours)}	                                & S+G & 81.9 & 91.0 & 93.0 & 82.3 & 85.2 & 90.6 & 93.1 & 88.1 & 96.1 & 89.7 & 87.7 & 95.2 & \textbf{86.8} & \textbf{88.0} & \textbf{93.7}  \\
\midrule
\multicolumn{17}{c}{\textbf{LesionSegAbdomen}}\\
\midrule
\multirow{2}{*}{Method} & \multirow{2}{*}{Type} &
\multicolumn{3}{c}{Colon} &
\multicolumn{3}{c}{Liver} &
\multicolumn{3}{c}{Pancreas} &
\multicolumn{3}{c}{Stomach} &
\multicolumn{3}{c}{Average} \\
\cmidrule(lr){3-5}\cmidrule(lr){6-8}\cmidrule(lr){9-11}\cmidrule(lr){12-14}\cmidrule(lr){15-17}
& & {SE} & {SP} & {AUC} & {SE} & {SP} & {AUC} & {SE} & {SP} & {AUC} &
    {SE} & {SP} & {AUC} & {SE} & {SP} & {AUC} \\
\midrule
nnUNet-c~\cite{nnUNet}              & S   & 92.0 & 93.0 & 96.0 & {--} & {--} & {--} & {--} & {--} & {--} & {--} & {--} & {--} & {--} & {--} & {--} \\
nnUNet-l~\cite{nnUNet}              & S   & {--} & {--} & {--} & 93.4 & 91.0 & 97.4 & {--} & {--} & {--} & {--} & {--} & {--} & {--} & {--} & {--} \\
nnUNet-p~\cite{nnUNet}              & S   & {--} & {--} & {--} & {--} & {--} & {--} & 85.4 & 88.0 & 94.0 & {--} & {--} & {--} & {--} & {--} & {--} \\
nnUNet-s~\cite{nnUNet}              & S   & {--} & {--} & {--} & {--} & {--} & {--} & {--} & {--} & {--} & 86.0 & 84.0 & 91.5 & {--} & {--} & {--} \\
nnUNet-fuse~\cite{nnUNet}           & S   & 92.0 & 93.0 & 96.0 & 93.4 & 91.0 & 97.4 & 85.4 & 88.0 & 94.0 & 86.0 & 84.0 & 91.5 & 89.2 & 89.0 & 94.7 \\
\midrule
CLIP~\cite{CLIP}                    & G   & 59.0 & 71.2 & 64.6 & 65.0 & 58.1 & 62.5 & 71.0 & 53.4 & 66.4 & 69.0 & 61.9 & 68.9 & 66.0 & 61.1 & 65.6 \\
LOVT~\cite{LOVT}                    & G   & 80.0 & 67.3 & 83.9 & 84.8 & 16.8 & 55.2 & 72.3 & 60.0 & 70.2 & 54.0 & 55.8 & 54.9 & 72.8 & 50.0 & 66.1 \\
Imitate~\cite{imitate}              & G   & 56.0 & 66.9 & 66.7 & 44.2 & 59.2 & 48.3 & 75.6 & 58.7 & 71.7 & 59.0 & 74.2 & 67.9 & 58.7 & 64.8 & 63.7 \\
BIUD~\cite{BIUD}                    & G   & 61.0 & 72.3 & 73.3 & 56.9 & 66.8 & 65.9 & 64.6 & 60.3 & 66.6 & 56.0 & 56.4 & 57.8 & 59.6 & 64.0 & 65.9 \\
Merlin~\cite{Merlin}                & G   & 60.0 & 47.3 & 52.9 & 56.3 & 52.1 & 55.2 & 67.6 & 53.0 & 64.8 & 44.0 & 67.0 & 53.3 & 57.0 & 54.9 & 56.5 \\
fVLM~\cite{fvlm}                    & G   & 80.0 & 75.5 & 82.9 & 61.9 & 64.8 & 65.9 & 70.3 & 68.6 & 77.4 & 71.0 & 76.6 & 75.7 & 70.8 & 71.4 & 75.5 \\
ViSD-Boost~\cite{VISD-Boost}        & G   & 67.0 & 71.0 & 76.9 & 65.0 & 71.4 & 73.6 & 82.0 & 80.1 & 88.7 & 61.0 & 64.7 & 67.1 & 68.7 & 71.8 & 76.6 \\
\midrule
OpenVocabCT~\cite{OpenVocabCT}      & S+G & 71.0 & 79.6 & 81.1 & 83.2 & 84.6 & 90.2 & 82.6 & 82.6 & 90.2 & 79.0 & 86.4 & 85.6 & 79.0 & 83.3 & 86.8  \\
HCFNet~\cite{HCFNet}                & S+G & 83.0 & 87.2 & 91.3 & 68.5 & 82.5 & 79.8 & 85.6 & 86.5 & 93.8 & 84.0 & 83.2 & 90.9 & 80.3 & 84.9 & 89.0  \\
VLWS~\cite{VLWS}                    & S+G & 77.0 & 81.2 & 86.0 & 77.2 & 85.6 & 86.8 & 81.2 & 77.9 & 87.5 & 80.0 & 83.1 & 88.0 & 78.9 & 81.9 & 87.0  \\
\midrule
\textbf{SuG (Ours)}	                                & S+G & 91.0 & 95.1 & 97.3 & 95.9 & 95.0 & 99.3 & 87.7 & 88.7 & 95.1 & 83.0 & 88.0 & 92.7 & \textbf{89.4} & \textbf{91.7} & \textbf{96.1}  \\
\bottomrule
\end{tabular}%
}
\caption{\textbf{Fine-grained lesion-centric diagnosis performance on four tumor categories (colon, liver, pancreas, and stomach) on the MedVL-CT69K and LesionSegAbdomen test sets.} ``--'' indicates that a specific category is not supported by the specialist model.}
\label{app_tab:medvl_specialist}
\end{table*}

\begin{table*}[t]
\centering
\caption{\textbf{Per-disease diagnostic performance on the MedVL-CT69K test set.} Reported metrics include sensitivity (SE), specificity (SP), accuracy (ACC), and AUC, all presented as percentages.}
\label{app_tab:medvl_detailed}
\scriptsize
\begin{tabular}{llcccc}
\toprule
Anatomy & Abnormality & SE & SP & ACC & AUC \\
\midrule
\multirow{2}{*}{Adrenal gland}
& Adrenal hypertrophy & 68.8 & 81.3 & 75.0 & 80.9 \\
& Nodule & 74.7 & 83.9 & 79.3 & 85.0 \\
\midrule
\multirow{2}{*}{Bladder}
& Diverticulum & 85.7 & 78.4 & 82.1 & 88.1 \\
& Stone & 92.9 & 90.5 & 91.7 & 96.0 \\
\midrule
\multirow{8}{*}{Colon}
& Colonic gas & 81.4 & 81.7 & 81.5 & 89.3 \\
& Effusion & 82.0 & 83.2 & 82.6 & 87.8 \\
& Obstruction & 94.1 & 98.3 & 96.2 & 99.1 \\
& Diverticulum & 72.1 & 75.6 & 73.8 & 80.5 \\
& Colon cancer & 86.5 & 90.8 & 88.6 & 93.9 \\
& Rectal cancer & 86.3 & 88.5 & 87.4 & 93.4 \\
& Appendicitis & 84.2 & 72.6 & 78.4 & 87.9 \\
& Appendicolith & 59.5 & 62.1 & 60.8 & 63.8 \\
\midrule
\multirow{1}{*}{Duodenum}
& Diverticulum & 86.7 & 93.3 & 90.0 & 94.9 \\
\midrule
\multirow{2}{*}{Esophagus}
& Hiatal hernia & 90.0 & 89.4 & 89.7 & 96.5 \\
& Varicose veins & 94.9 & 97.2 & 96.0 & 98.0 \\
\midrule
\multirow{3}{*}{Gallbladder}
& Inflammation & 83.7 & 75.4 & 79.5 & 87.0 \\
& Gallstone & 87.0 & 89.8 & 88.4 & 94.3 \\
& Adenomyomatosis & 75.0 & 62.5 & 68.8 & 73.3 \\
\midrule
\multirow{2}{*}{Heart}
& Cardiomegaly & 95.0 & 90.1 & 92.5 & 96.8 \\
& Pericardial effusion & 80.5 & 78.8 & 79.7 & 86.5 \\
\midrule
\multirow{4}{*}{Kidney}
& Atrophy & 86.5 & 79.3 & 82.9 & 87.4 \\
& Cyst & 76.2 & 81.9 & 79.0 & 86.1 \\
& Hydronephrosis & 86.2 & 86.5 & 86.3 & 93.9 \\
& Renal calculus & 71.8 & 84.5 & 78.2 & 82.2 \\
\midrule
\multirow{8}{*}{Liver}
& Fatty liver & 92.4 & 90.6 & 91.5 & 96.5 \\
& Glisson's capsule effusion & 95.6 & 86.6 & 91.1 & 93.9 \\
& Metastatic tumor & 86.1 & 90.9 & 88.5 & 93.7 \\
& Intrahepatic bile duct dilatation & 76.9 & 78.4 & 77.6 & 86.3 \\
& Cancer & 93.4 & 95.3 & 94.3 & 98.5 \\
& Cyst & 81.8 & 79.0 & 80.4 & 87.4 \\
& Abscess & 83.3 & 80.0 & 81.7 & 83.8 \\
& Cirrhosis & 89.9 & 94.2 & 92.1 & 97.3 \\
\midrule
\multirow{5}{*}{Lung}
& Atelectasis & 94.3 & 98.4 & 96.4 & 98.1 \\
& Bronchiectasis & 88.9 & 98.4 & 93.7 & 96.2 \\
& Emphysema & 80.0 & 63.9 & 72.0 & 75.4 \\
& Pneumonia & 82.8 & 74.2 & 78.5 & 87.3 \\
& Pleural effusion & 94.3 & 97.5 & 95.9 & 97.6 \\
\midrule
\multirow{5}{*}{Pancreas}
& Pancreatic cancer & 86.2 & 85.2 & 85.7 & 89.0 \\
& Atrophy & 81.1 & 84.0 & 82.5 & 88.6 \\
& Pancreatitis & 87.0 & 95.4 & 91.2 & 95.8 \\
& Pancreatic duct dilatation & 85.1 & 80.8 & 82.9 & 90.6 \\
& Steatosis & 82.2 & 83.2 & 82.7 & 87.8 \\
\midrule
\multirow{2}{*}{Portal vein}
& Hypertension & 94.4 & 95.2 & 94.8 & 98.7 \\
& Thrombosis & 94.5 & 92.6 & 93.6 & 98.1 \\
\midrule
\multirow{1}{*}{Sacrum}
& Osteitis & 88.2 & 80.3 & 84.3 & 88.3 \\
\midrule
\multirow{4}{*}{Small intestine}
& Gas accumulation & 84.0 & 76.3 & 80.2 & 86.5 \\
& Fluid accumulation & 88.7 & 85.3 & 87.0 & 92.5 \\
& Obstruction & 93.4 & 91.6 & 92.5 & 95.7 \\
& Intussusception & 66.7 & 75.6 & 71.1 & 74.9 \\
\midrule
\multirow{3}{*}{Spleen}
& Hemangioma & 87.2 & 92.1 & 89.7 & 94.2 \\
& Infarction & 100.0 & 93.4 & 96.7 & 98.7 \\
& Splenomegaly & 87.5 & 85.8 & 86.7 & 93.8 \\
\midrule
\multirow{2}{*}{Stomach}
& Gastric wall thickening & 74.8 & 80.8 & 77.8 & 84.4 \\
& Gastric cancer & 86.3 & 86.0 & 86.2 & 92.5 \\
\bottomrule
\end{tabular}
\end{table*}

\begin{table}[t]
\centering
\caption{\textbf{Per-disease diagnostic performance on the CT-RATE test set.} Reported metrics include precision (Prec), accuracy (ACC), F1, and ROC AUC (AUC), all presented as percentages.}
\label{tab:ctrate_detailed}
\begin{tabular}{lcccc}
\toprule
Abnormality & Prec & ACC & F1 & AUC \\
\midrule
Emphysema & 34.8 & 68.9 & 71.8 & 75.1 \\
Atelectasis & 35.1 & 63.8 & 66.4 & 69.4 \\
Lung nodule & 56.9 & 61.7 & 61.8 & 65.6 \\
Lung opacity & 60.8 & 70.7 & 71.0 & 77.8 \\
Pulmonary fibrotic sequela & 33.9 & 54.8 & 57.1 & 62.1 \\
Pleural effusion & 66.8 & 93.4 & 93.9 & 96.7 \\
Mosaic attenuation pattern & 16.2 & 66.1 & 73.7 & 75.8 \\
Peribronchial thickening & 28.2 & 75.7 & 79.6 & 80.3 \\
Consolidation & 47.7 & 79.2 & 81.1 & 88.3 \\
Bronchiectasis & 22.1 & 72.1 & 77.0 & 72.9 \\
Interlobular septal thickening & 22.7 & 75.7 & 81.0 & 86.4 \\
Cardiomegaly & 39.1 & 84.5 & 86.8 & 93.8 \\
Pericardial effusion & 25.2 & 81.6 & 85.5 & 86.8 \\
Coronary artery wall calcification & 65.7 & 85.3 & 85.9 & 92.6 \\
Hiatal hernia & 28.5 & 70.4 & 74.8 & 79.9 \\
Arterial wall calcification & 69.6 & 86.1 & 86.5 & 93.7 \\
\bottomrule
\end{tabular}
\end{table}

\begin{table}[t]
\centering
\caption{\textbf{Per-disease diagnostic performance on the RAD-ChestCT test set.} Reported metrics include precision (Prec), accuracy (ACC), F1, and ROC AUC (AUC), all presented as percentages.}
\label{tab:radchest_detailed}
\begin{tabular}{lcccc}
\toprule
Abnormality & Prec & ACC & F1 & AUC \\
\midrule
Emphysema & 45.9 & 69.3 & 70.8 & 74.6 \\
Atelectasis & 36.8 & 57.1 & 59.0 & 60.1 \\
Lung nodule & 81.8 & 61.7 & 64.3 & 63.7 \\
Lung opacity & 58.8 & 56.5 & 56.5 & 58.0 \\
Pulmonary fibrotic sequela & 29.2 & 73.4 & 77.4 & 80.5 \\
Pleural effusion & 60.5 & 85.7 & 86.5 & 92.2 \\
Peribronchial thickening & 12.6 & 56.7 & 66.0 & 64.2 \\
Consolidation & 28.6 & 71.4 & 75.6 & 78.8 \\
Bronchiectasis & 26.1 & 63.9 & 68.7 & 71.0 \\
Interlobular septal thickening & 15.1 & 69.2 & 76.8 & 78.1 \\
Cardiomegaly & 28.6 & 76.1 & 80.3 & 85.3 \\
Pericardial effusion & 26.3 & 67.6 & 71.8 & 68.9 \\
Hiatal hernia & 29.1 & 77.0 & 80.5 & 78.2 \\
\bottomrule
\end{tabular}
\end{table}

\begin{table}[t]
\centering
\caption{\textbf{Per-disease diagnostic performance on the Merlin test set.} Reported metrics include F1, ROC AUC (AUC), and accuracy (ACC), all presented as percentages.}
\label{tab:merlin_detailed}
\begin{tabular}{lccc}
\toprule
Abnormality & F1 & AUC & ACC \\
\midrule
aortic\_valve\_calcification & 95.7 & 98.7 & 95.7 \\
pleural\_effusion & 92.7 & 97.5 & 92.7 \\
splenomegaly & 91.0 & 97.2 & 91.1 \\
bowel\_obstruction & 93.6 & 98.1 & 93.8 \\
abdominal\_aortic\_aneurysm & 94.1 & 94.8 & 94.1 \\
hepatomegaly & 86.0 & 92.3 & 86.8 \\
pancreatic\_atrophy & 85.2 & 91.1 & 84.8 \\
hepatic\_steatosis & 88.5 & 93.0 & 88.9 \\
surgically\_absent\_gallbladder & 90.0 & 90.5 & 90.5 \\
cardiomegaly & 86.1 & 91.9 & 85.7 \\
hydronephrosis & 85.6 & 90.6 & 85.7 \\
atherosclerosis & 83.9 & 89.7 & 83.6 \\
submucosal\_edema & 77.5 & 85.0 & 77.4 \\
hiatal\_hernia & 80.3 & 87.6 & 80.1 \\
biliary\_ductal\_dilation & 82.5 & 87.3 & 80.8 \\
renal\_cyst & 75.9 & 82.9 & 75.5 \\
atelectasis & 79.0 & 86.9 & 79.1 \\
gallstones & 68.7 & 75.7 & 69.3 \\
renal\_hypodensities & 68.2 & 73.5 & 66.0 \\
fracture & 71.6 & 76.2 & 71.8 \\
appendicitis & 51.4 & 52.1 & 58.1 \\
osteopenia & 83.5 & 88.9 & 83.3 \\
ascites & 84.5 & 92.2 & 84.2 \\
free\_air & 86.3 & 92.4 & 85.6 \\
coronary\_calcification & 76.9 & 81.0 & 76.2 \\
anasarca & 80.2 & 87.3 & 80.1 \\
lymphadenopathy & 71.0 & 73.4 & 68.4 \\
metastatic\_disease & 76.8 & 80.4 & 75.0 \\
prostatomegaly & 74.6 & 75.3 & 73.5 \\
thrombosis & 62.0 & 53.4 & 57.8 \\
\bottomrule
\end{tabular}
\end{table}

\begin{table}[t]
\centering
\caption{\textbf{Per-disease \textit{unbalanced} diagnostic performance on the Merlin test set.} Reported metrics include F1, ROC AUC (AUC), and accuracy (ACC), all presented as percentages.}
\label{tab:merlin_unbalanced_detailed}
\begin{tabular}{lccc}
\toprule
Abnormality & F1 & AUC & ACC \\
\midrule
aortic\_valve\_calcification & 95.1 & 98.1 & 95.0 \\
pleural\_effusion & 93.4 & 97.3 & 92.2 \\
splenomegaly & 40.1 & 96.7 & 89.1 \\
abdominal\_aortic\_aneurysm & 80.0 & 95.9 & 92.6 \\
bowel\_obstruction & 42.9 & 95.0 & 89.0 \\
pancreatic\_atrophy & 31.2 & 91.0 & 84.4 \\
hepatomegaly & 38.8 & 88.8 & 83.8 \\
biliary\_ductal\_dilation & 63.5 & 89.5 & 83.6 \\
cardiomegaly & 77.8 & 92.3 & 86.2 \\
hepatic\_steatosis & 92.8 & 92.5 & 86.5 \\
surgically\_absent\_gallbladder & 86.1 & 90.2 & 90.2 \\
submucosal\_edema & 76.7 & 85.2 & 77.8 \\
hydronephrosis & 31.9 & 90.1 & 83.1 \\
atherosclerosis & 91.1 & 89.9 & 83.2 \\
renal\_cyst & 63.0 & 83.2 & 75.7 \\
hiatal\_hernia & 55.0 & 88.3 & 80.9 \\
atelectasis & 78.2 & 86.7 & 79.0 \\
renal\_hypodensities & 26.2 & 75.5 & 69.2 \\
gallstones & 62.3 & 75.3 & 69.1 \\
fracture & 71.4 & 77.2 & 72.8 \\
appendicitis & 47.5 & 49.8 & 54.8 \\
osteopenia & 54.9 & 90.5 & 83.7 \\
ascites & 89.0 & 92.2 & 84.5 \\
free\_air & 26.5 & 87.3 & 78.2 \\
coronary\_calcification & 43.5 & 80.8 & 74.0 \\
anasarca & 54.7 & 87.7 & 81.7 \\
metastatic\_disease & 62.9 & 81.2 & 75.3 \\
lymphadenopathy & 47.2 & 75.1 & 70.7 \\
prostatomegaly & 23.9 & 74.6 & 72.2 \\
thrombosis & 73.0 & 53.7 & 58.5 \\
\bottomrule
\end{tabular}
\end{table}

% \clearpage
% \FloatBarrier
% \input{checklist.tex}

\end{document}